\documentclass{article}
    \PassOptionsToPackage{numbers, compress}{natbib}

    \usepackage[final]{neurips_2024}

\usepackage[utf8]{inputenc} 
\usepackage[T1]{fontenc}    
\usepackage{hyperref}       
\usepackage{url}            
\usepackage{booktabs}       
\usepackage{amsfonts}       
\usepackage{nicefrac}       
\usepackage{microtype}      
\usepackage{xcolor}         
\usepackage{dsfont}
\usepackage{graphicx}
\usepackage{fancybox}
\usepackage{amsmath}
\usepackage{multirow}
\definecolor{attributeloss}{RGB}{184, 84, 80}
\definecolor{matchingloss}{RGB}{08, 142, 191}
\definecolor{VQ}{RGB}{214,182,86}
\definecolor{CL}{RGB}{114,82,86}
\definecolor{RE}{RGB}{150,115,166}

\usepackage[ruled]{algorithm2e}
\title{MO-DDN: A Coarse-to-Fine Attribute-based Exploration Agent for Multi-object Demand-driven Navigation}

\author{
	Hongcheng Wang$^{1,3}$\footnotemark[1] \quad Peiqi Liu$^{2}$\thanks{Equal contribution.} \\ \textbf{Wenzhe Cai}$^{4}$\quad \textbf{Mingdong Wu} $^{1,3}$ \quad \textbf{Zhengyu Qian} $^{2}$\quad \textbf{Hao Dong} $^{1,3}$\thanks{Corresponding author.}\\
  $^1$CFCS, School of CS, PKU \quad $^2$School of EECS, PKU \quad $^3$PKU-Agibot Lab \quad \\$^4$School of Automation, Southeast University
}

\begin{document}

\maketitle

\begin{abstract}
The process of satisfying daily demands is a fundamental aspect of humans' daily lives. With the advancement of embodied AI, robots are increasingly capable of satisfying human demands. Demand-driven navigation (DDN) is a task in which an agent must locate an object to satisfy a specified demand instruction, such as ``I am thirsty.'' The previous study typically assumes that each demand instruction requires only one object to be fulfilled and does not consider individual preferences. However, the realistic human demand may involve multiple objects. In this paper, we introduce the Multi-object Demand-driven Navigation (MO-DDN) benchmark, which addresses these nuanced aspects, including multi-object search and personal preferences, thus making the MO-DDN task more reflective of real-life scenarios compared to DDN. Building upon previous work, we employ the concept of ``attribute'' to tackle this new task. However, instead of solely relying on attribute features in an end-to-end manner like DDN, we propose a modular method that involves constructing a coarse-to-fine attribute-based exploration agent (C2FAgent). Our experimental results illustrate that this coarse-to-fine exploration strategy capitalizes on the advantages of attributes at various decision-making levels, resulting in superior performance compared to baseline methods. Code and video can be found at \href{https://sites.google.com/view/moddn}{https://sites.google.com/view/moddn}.
\end{abstract}
\begin{figure}[h]
  \centering
  \includegraphics[width=1\textwidth,trim=00 00 00 00,clip]{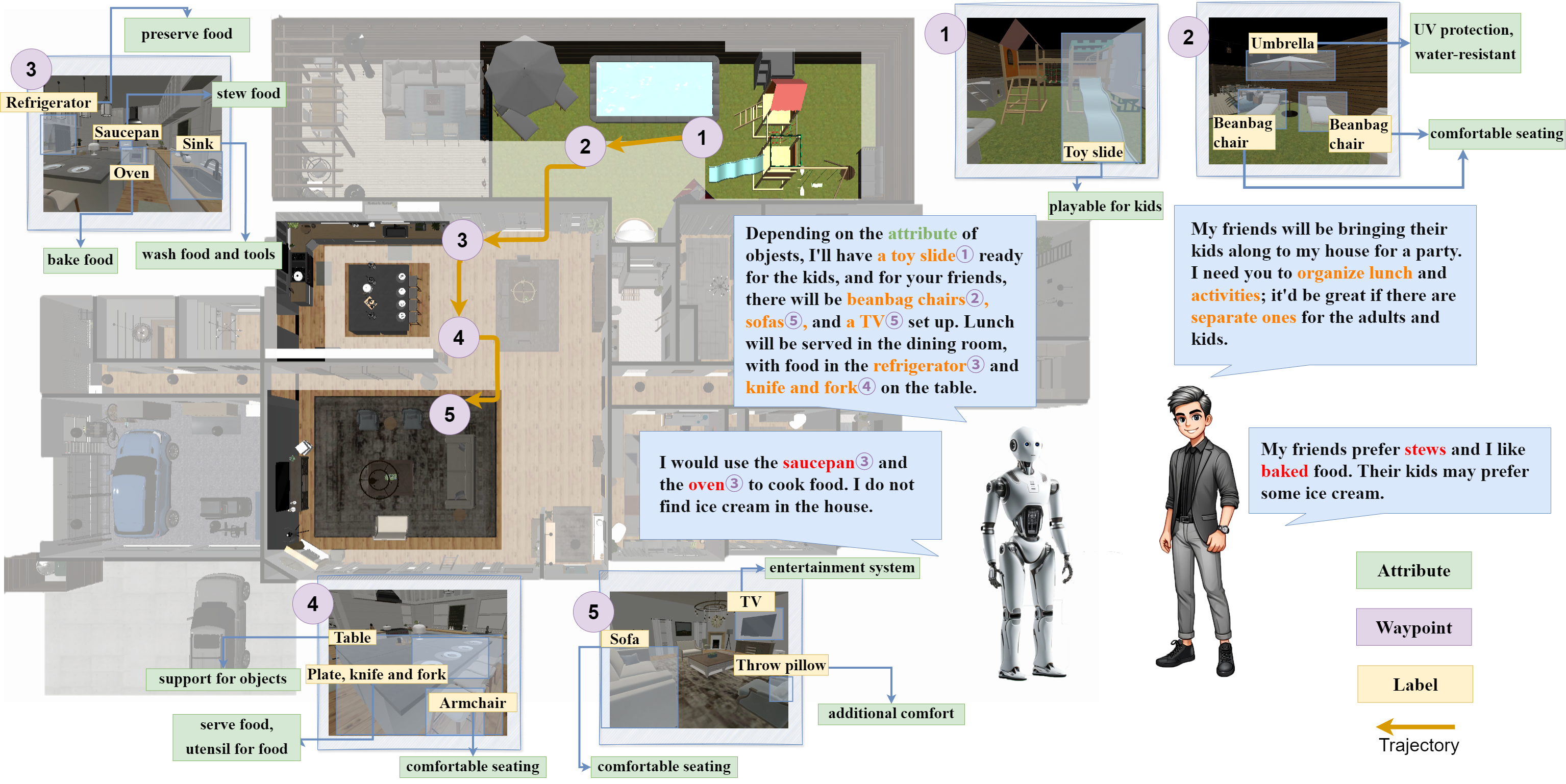}
  \caption{\textbf{An example of Multi-object Demand-driven Navigation.} A user plans to host a party at his new house and outlines some basic demands (\textcolor{orange}{highlighted in orange}), along with specific preferences for different individuals (\textcolor{red}{highlighted in red}). The agent parses these demands and locates multiple objects in various locations in the scene to fulfill them. Despite not meeting the preferred ``ice cream'' demand, the agent successfully addresses basic demands, such as organizing lunch.}
  \label{fig:teaser}
\end{figure}

\section{Introduction}
  In the field of psychology, the creation of demands directs the motivation for human behavior in the real world, and behavior ultimately leads to the fulfillment of the demand~\citep{taormina2013maslow, gawel2019herzberg,maslow1981motivation,maslow1958dynamic}. For example, an individual might crave sweet iced tea, prompting them to search for object such as tea, sugar, ice, and water, and then combine them to make sweet iced tea.
  Recently, with the rapid development of embodied AI and large language models, researchers are interested in using robots to meet various human demands~\citep{ahn2022can,liang2023code,huang2022inner,brohan2022rt,driess2023palm,brohan2023rt, huang2023voxposer,wu2022vat,zhou2024navgpt}. Demand-driven Navigation (DDN)~\citep{wang2024find} is a variant of ObjectGoal Navigation (ON)~\citep{du2021vtnet,cai2023bridging,chaplot2020object,zhu2021soon,fukushima2022object,zhou2023esc,majumdar2022zson}, which requires an agent to find an object that satisfies a given demand instruction. For example, when a user gives the agent an instruction such as ``I am thirsty,'' the robot has to search the entire environment for objects such as water, tea, coffee, etc., depending on what is available within the environment.

  In previous work~\citep{wang2024find}, demand instructions often exhibit a low level of complexity, lack consideration for user preferences, and typically require only one object to fulfill each instruction. However, real-world situations frequently involve more intricate instructions, necessitating the coordination of multiple objects to satisfy a demand. For example, in the BEHAVIOR-1K~\citep{li2023behavior}, daily tasks often involve the interaction of multiple objects.
  Furthermore, individual users may have distinct preferences. For instance, when presented with the basic demand ``I am thirsty,'' one user may prefer sparkling water while another may favor juice. 
  However, satisfying preferred demands is obviously more difficult than satisfying basic demands, so the agent needs to be flexible in prioritizing goals depending on both the situations of users and environments. 

  We introduce a new benchmark, Multi-object Demand-driven Navigation (MO-DDN), in which demand instructions are comprised of basic and preferred demand components. An agent must find a combination of objects (we call it a ``\textbf{solution}'' later) that satisfies the demand. The agent should satisfy the user's preferred demands as much as the situation allows.
  We use GPT-4 \footnote{we use gpt-4-0125-preview API in this paper here and later, unless otherwise noted.} to automatically generate and modify tasks and perform manual checks. 
  Our settings are similar to Multi-object Navigation (MON)~\citep{marza2023multi,wani2020multion,chen2022learning}, but we use a demand instruction to combine potentially multiple objects in a form that is consistent with common sense and personal preferences of humans rather than a list of object categories in MON. The number and category of objects in a solution are not known in advance but require an agent to reason on its own. Moreover, solutions that satisfy the same demand may have a different number of objects. We argue that MO-DDN has similar advantages over MON as DDN has over ON.

  We consider MO-DDN to be a crucial preliminary step in task planning. Today, with sufficient meta-information provided (\emph{e.g.}, position, state, and category of all objects in the scene), large language models can provide satisfactory task planning results~\citep{liu2024egocentric,lin2023grounded,wu2023embodied,mendez2023embodied,ahn2022can}.
  \textbf{However, maintaining and providing accurate meta-information can be challenging and inconvenient in real life.} The user's interactions with the scene change the meta-information frequently.  Moreover, since the user is not necessarily omniscient about the scene, especially in unfamiliar scenes, it is possible that if the user parses the demand himself and directly provides the desired objects, these desired objects may not exist in the scene. Even if the user can provide comprehensive meta-information, delivering the meta-information to the agent is time-consuming and inconvenient for the user. 
  Therefore, the purpose of MO-DDN is to provide timely and accurate meta-information depending on \textbf{a convenient demand instruction} in natural language for later task planning or, more later, object manipulation.

  In the recently ObjectGoal Navigation work, benefiting from the development of Vision Language Models (VLMs)~\citep{Cheng2024YOLOWorld, wu2023general,alayrac2022flamingo,yang2023dawn, chen2023minigpt,zhu2023minigpt,wang2024visionllm} and Large Language Models (LLMs)~\citep{zhao2023survey,chowdhery2023palm, brown2020language,touvron2023llama2,shen2024hugginggpt,team2023gemini}, some ObjectGoal Navigation methods return to a modular manner~\citep{raychaudhuri2024mopa,gupta2017cognitive, chaplot2020object,luo2022stubborn,rudra2023contextual,liang2021sscnav,georgakis2021learning,min2023self,kumar2021gcexp} rather than an end-to-end manner~\citep{fukushima2022object,shen2019situational,mousavian2019visual,druon2020visual,yang2018visual,pal2021learning,mayo2021visual,dang2022unbiased,ye2021auxiliary}. While maintaining the core concept of the ``attribute'' from the end-to-end agent of the previous work~\citep{wang2024find}, we modify the training of the attribute model to make it applicable to multi-object search, and propose a coarse-to-fine attribute-based exploration modular agent, \textbf{C2FAgent}. In the coarse exploration phase, similar to the modular methods in ObjectGoal Navigation, we use a depth camera to reconstruct the environment point clouds and an object detection module, GLEE~\citep{wu2023general}, to label the objects in the point clouds. Additionally, the point clouds are compressed and segmented into several 2D rectangular blocks, and thus each detected object belongs to a block. For each block, we calculate the similarity of the objects' attribute features and the instruction's attribute features as a metric for choosing a waypoint. We calculate basic demand similarity scores and preferred demand similarity scores separately and select blocks according to a weighted sum of the two scores. 
  In the fine exploration phase (\emph{i.e.}, when the agent arrives at the waypoint), we train an end-to-end attribute exploration module to identify and report the target object to compose the solution. 
  
  The experimental results show that our proposed method outperforms the baselines, and the ablation study shows that the attribute model improves exploration efficiency in different phases.  We argue that this coarse-to-fine design allows for the incorporation of prior knowledge from external foundation models in the coarse exploration phase and task-relevant world-grounding exploration in the fine exploration phase. Moreover, the ablation study on the weighted sum of the basic and preferred similarity scores demonstrates that increasing the preferred weights allows the agent to prioritize searching for the preferred solution and increasing the basic weights allows the agent to prioritize searching for the basic solutions. Therefore, users can adjust the weights to influence the agent's behavior freely, which is a flexible way to handle personal preferences.
  
  Our main contributions are listed as follows:
  \begin{itemize}
      \item We propose a new benchmark, MO-DDN, which considers multi-object combinations as solutions and more complex and diverse demand instructions. MO-DDN can be regarded as a crucial preliminary step in task planning.

      \item We extend the training process of the attribute model, enabling the attribute features to work well in a multi-object setting. Based on the new version of attribute features, We design a coarse-to-fine attribute-based exploration agent, C2FAgent, for this benchmark, allowing the attribute features to play an important role in different exploration phases.

      \item The experimental results show that the attribute features do improve the efficiency of exploration, and the experimental results substantially surpass the baselines. Ablation study shows that attribute-based exploration is more efficient than frontier-based exploration~\citep{yamauchi1997frontier} and LLM-based waypoint selection. 
  \end{itemize}

\section{Related Work}
\subsection{Visual Navigation}
\paragraph{Goal Description} In general terms, visual navigation involves the continuous generation of actions based on RGB-D and GPS+Compass inputs until a specified objective is reached~\cite{anderson2018evaluation}. Visual navigation tasks vary in their goal description, such as step-by-step instructions in Vision-Language Navigation (VLN)~\cite{anderson2018vision, zhu2020vision,wang2021structured,majumdar2020improving,wang2020vision,long2023discuss,hong2021vln,shah2023lm}, audio in Audio-visual Navigation~\cite{chen2021semantic, chen2020soundspaces,chen2020learning,wang2023learning,yu2022sound,younes2023catch}, object categories in ObjectGoal Navigation (ON)~\cite{du2021vtnet,cai2023bridging,chaplot2020object,zhu2021soon,fukushima2022object,zhou2023esc,majumdar2022zson}, object category lists in MultiObject Navigation (MON)~\cite{raychaudhuri2024mopa,marza2023multi,wani2020multion,chen2022learning}, and demand instructions in Demand-driven Navigation (DDN)~\cite{wang2024find}. Our proposed benchmark, MO-DDN, can be viewed as a multi-object version of DDN. Although, like DDN, MO-DDN's inputs are demand instructions,  MO-DDN's solutions are multi-object rather than single-object, as in DDN.

\paragraph{Previous Method Overview} In ON, with the rise of object detection models~\citep{wu2023general,Cheng2024YOLOWorld,hussain2023yolo,carion2020end}, object segmentation models~\citep{kirillov2023segment, ke2024segment,ren2024segment,o2015learning}, and large language models, modular methods~\citep{raychaudhuri2024mopa,gupta2017cognitive, chaplot2020object,luo2022stubborn,rudra2023contextual,liang2021sscnav,georgakis2021learning,min2023self,kumar2021gcexp} are gradually being developed. They greatly improve navigation efficiency and success rate by building semantic maps and navigable maps and then planning paths on the maps. Meanwhile, end-to-end methods~\citep{fukushima2022object,shen2019situational,mousavian2019visual,druon2020visual,yang2018visual,pal2021learning,mayo2021visual,dang2022unbiased,ye2021auxiliary} focus more on learning associations in object-object and object-scene to help reason about the potential location of target objects.
In MON, previous work has focused on studying how to quickly construct~\citep{chen2022learning}, memorize~\citep{wani2020multion,raychaudhuri2024mopa}, and use semantic maps ~\citep{kim2021sgolam} or implicit representations~\citep{marza2023multi} of scenes. In DDN, the concept of attributes is introduced as a way of expressing what an object shows when it fulfills a demand, \emph{e.g.}, quenching thirst is an attribute of water in the context of demand ``I am thirsty''. 
In this paper, we extend the concept of attributes and propose a new method to train the attribute model. We combine the advantages of modular and end-to-end methods and propose a coarse-to-fine attribute-based exploration modular agent.

\subsection{Foundation Large Model in Embodied Task}
Foundation Large Models refer to self-supervised pre-trained models trained on large Internet-scale datasets, which have demonstrated promising capabilities across various embodied tasks. Large language models (LLMs) such as GPT-4~\citep{achiam2023gpt}, LLaMA~\citep{touvron2023llama,touvron2023llama2}, and Gemma~\citep{team2024gemma} exhibit performance comparable to humans in task planning~\citep{liu2024egocentric,lin2023grounded,wu2023embodied,mendez2023embodied,song2023llmplanner}, common sense reasoning~\citep{zhao2024large} and question answering~\citep{shao2023prompting, yu2024self,guo2023images,cui2023chatlaw,chen2023minigpt,zhuang2024toolqa}. Furthermore, researchers use LLMs to synthesize task datasets~\citep{wang2024find}. The CLIP model~\citep{radford2021learning}, which employs contrastive learning with image and text pairs, showcases strong semantic extraction abilities in navigation~\citep{wang2024find,dorbala2022clip, majumdar2022zson,shah2023lm,khandelwal2022simple}. Visual Language Models offer understanding of images such as image descriptions~\citep{kinghorn2018region, vinyals2015show,chen2015mind,wang2023caption}, object detection~\citep{wu2023general,Cheng2024YOLOWorld,hussain2023yolo,carion2020end}, and object segmentation~\citep{kirillov2023segment, ke2024segment,ren2024segment,o2015learning} for downstream embodied tasks.
This paper uses LLMs, specifically GPT-4, to generate task instructions, solutions, and attribute examples for attribute learning. Furthermore, we employ GLEE~\citep{wu2023general}, a state-of-the-art object detection model, to detect object categories within the field of view for attribute feature extraction.

\section{Multi-object Demand-driven Navigation}
\label{moddn_defi}
Following the basic settings in DDN, let $\mathcal{D}$ denote a set of demand instructions, $\mathcal{S}e$ denotes a set of navigable scenes, and $\mathcal{O}$ denotes a set of object categories. Let $\mathcal{S}o$ denote a set of solutions for demand instructions. An element in $\mathcal{S}o$ (\emph{i.e.}, a solution) is a subset of $\mathcal{O}$.
In each episode, similar to the DDN task, an agent is randomly initialized to a location within a mapless environment and receives a demand instruction $DI \in \mathcal{D}$ in natural language. In this benchmark, a demand instruction consists of two parts: one is basic demand instruction $DI_b$, and the other is preferred demand instruction $DI_p$, \emph{e.g.}, `` I need a comfortable place to play computer games, preferably with good lighting.'' Each $DI$ has two sets of solutions, \emph{i.e.}, basic solution $So_b$ and preferred solution $So_p$ for $DI_b$ and $DI_p$, respectively.  For example, \{desk, soft chair\} is an element in $So_b$ for the above example demand instruction, and \{desk, soft chair, table lamp\} is an element in $So_p$.

Then, at each time step, the agent should choose an action from $\mathrm{MoveAhead}$, $\mathrm{RotateRight}$, $\mathrm{RotateLeft}$, $\mathrm{LookUp}$, $\mathrm{LookDown}$, $\mathrm{Find}$, and $\mathrm{Done}$. 
The action $\mathrm{Find}$ is similar in MON, which automatically reports the objects in the field of view. To reduce the difficulty of the MO-DDN task and focus it on navigation, if the agent chooses $\mathrm{Find}$, all objects in the field of view with distance below the threshold $d_{find}$ will be recorded in a found list $FL$, instead of requiring the bounding box of the target object as in DDN. The found list $FL$ is used to calculate the basic and preferred success rate later.
When the agent chooses $\mathrm{Done}$, or the number of choices $\mathrm{Find}$ reaches a threshold $n_{find}$, or the number of steps reaches a threshold $n_{step}$, the episode ends and the success rate is calculated. 
For a specific demand instruction $DI$, We calculate the basic success rate as follows:
\begin{equation}
    SR_{b}=\frac{1}{N} \sum^N_{i=1} \max_{s_b \in So_b} \frac{\sum_{o \in FL} \mathds{1}_{o \in s_b}}{Len(s_b)}
\end{equation}
where $N$ donates the number of testing episodes, $s_b$ donates a basic solution in the solution set of $DI$, $FL$ donates the found list that the agent reports, $o$ donates an object category. The preferred success rate $SR_{p}$ is calculated similarly. To summarize, the success rate of an episode is the value with the highest percentage of satisfaction among all solutions. We also calculate the SPL~\cite{anderson2018evaluation} corresponding to $SR_{b}$ and $SR_{p}$. We generate 300 tasks, encompassing 358 object categories from the HSSD dataset~\cite{khanna2023hssd}. 
These tasks are referred to as HSSD's world-grounding tasks, indicating that objects in these tasks are in the HSSD dataset. We generate some language-grounding tasks to train the attribute model afterward. Language-grounding means that the objects in the solutions can be everything that makes sense rather than restricting objects in HSSD.
Please see the supplementary material for details about task generation~\ref{task_generation}, task metrics~\ref{srspl}, and task dataset statistics~\ref{task_statistics}.

\section{Method}
\subsection{Attribute Model Training}
\label{attribute_feature_training}
In this section, we describe how to train the MO-DDN's attribute model that extends from the attribute model in DDN. In DDN, the training of attribute features is constrained by the assumption that each instruction requires only one attribute that can be satisfied by a single object. Such attribute features may not work well under multi-object settings. This paper proposes directly mapping demand instructions and object categories into the same attribute feature space. Such a mapping can learn multiple attribute features simultaneously to address multi-object search. 
Concretely, the core function of the attribute model is to map a demand instruction in $\mathcal{D}$ or an object category in $\mathcal{O}$ to several attribute features in $\mathcal{R}^{d}$ ( $\mathcal{R}$ is the set of real numbers, $d$ is the dimension of the attribute features) that are in the shared attribute feature space. In order to enable the alignment of the attribute features of instructions and objects, we design a discrete codebook and five losses. The alignment means that for a demand instruction and an object in its solution, one of the attribute features of the instruction and one of the attribute features of the object have a high cosine similarity.

\begin{figure}[ht]
  \centering
  \includegraphics[width=\textwidth,trim=00 00 00 00,clip]{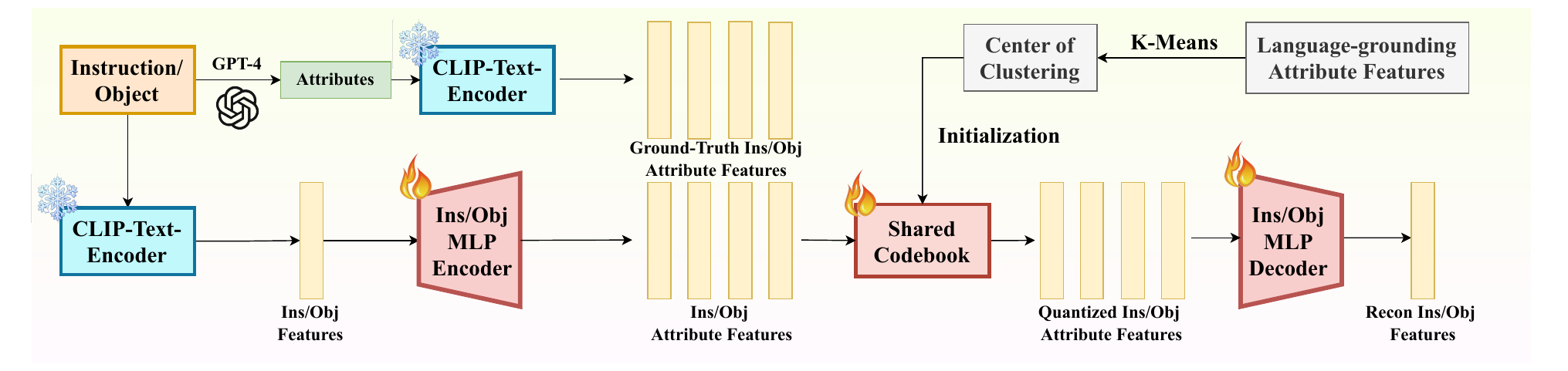}
  \caption{\textbf{Attribute Model.} This figure shows the architecture of the attribute model. Instructions and objects share the same model architecture. Instructions and items share the same model architecture. For parameters, they share only the parameters of the shared codebook, while the parameters of the MLP Encoder and Decoder are independent. Only the red with flames modules in the figure will be trained while the blue with snowflakes CLIP model parameters will be frozen. 
  }
  \label{fig:attribute}
\end{figure}
\subsubsection{Codebook and Its Initialization}

The reason for using a discrete codebook is similar to VQ-VAE, \emph{i.e.}, different attributes are inherently discrete from each other. Then using a discrete codebook as an attribute feature space is a better way to represent the relationships between attributes than a continuous attribute feature space.
The codebook is essentially some feature vectors, and in our experiments, we choose 128 as the number of vectors and 768 as the vector dimension (same with CLIP ViT-L/14's text dimension); thus, the codebook is a 128$\times$768 matrix.
we use CLIP-Text-Encoder (ViT-L/14) to encode the language-grounding attributes generated in the supplementary material Sec~\ref{lgtask} into attribute features. These attribute features are then clustered using K-means~\citep{lloyd1982least} to get 128 clustering centers. The feature vectors of these 128 clustering centers are used to initialize the codebook, making the codebook as a subspace of the CLIP feature space.

\subsubsection{Definition of Losses}
In our method, there are two attribute models that share the same architecture, the instruction attribute model $AM_{ins}$ and the object attribute model $AM_{obj}$, shown in Fig.~\ref{fig:attribute}. In both models, only MLP Encoder and CLIP Encoder will be used in \textbf{C2FAgent}, Codebook and MLP Decoder are only used to train MLP Encoder.
We obtain the $k_1$ instruction attributes and $k_2$ object attributes (the green square in Fig.~\ref{fig:attribute}) by prompting GPT-4 for ``what attributes can satisfy this instruction?'' and ``what attributes does this object have?'', respectively. 
For the five losses used to train the two attribute models, see the pseudo-code~\ref{pseudo_attribute}. 

\begin{algorithm}[H]
\small 
\label{pseudo_attribute}
    \SetAlgoLined 
    \SetNlSty{}{\hspace{2cm}}{}
	\caption{Losses in Attribute Training}
	\KwIn{An instruction and an object in the intruction's solution}
	\KwOut{Five losses}
    \Begin{
    \tcp{Variables in parentheses are full names.} 
    1. Ins Attributes = AskGPT-4(Instruction)\; 
    2. Obj Attributes = AskGPT-4(Objects) \;
    3. GT-IAF (GT Ins Attribute Features)= CLIP(Ins Attributes) \tcp{shape=($k_1$, $d$)} 
    4. GT-OAF (GT Obj Attribute Features)= CLIP(Obj Attributes) \tcp{shape=($k_2$, $d$)} 
    5. IF (Ins Feature) = CLIP(Instruction) \tcp{shape=(1, $d$)} 
    6. OF (Obj Feature) = CLIP(Objects) \tcp{shape=(1, $d$)} 
    7. IAF (Ins Attribute Features) = Ins MLP\_Encoder(IF) \tcp{shape=($k_1$, $d$)} 
    8. OAF (Obj Attribute Features)= Obj MLP\_Encoder(OF) \tcp{shape=($k_2$, $d$)}
    9. \textcolor{attributeloss}{Attribute Loss} = $MSE$(GT-IAF, IAF) + $MSE$(GT-OAF, OAF) \;
    10. Q-IAF (Quantized IAF) = Codebook(IAF)\tcp{shape=($k_1$, $d$)} 
    11. Q-OAF (Quantized OAF) = Codebook(OAF)\tcp{shape=($k_2$, $d$)}
    12. \textcolor{CL}{Commitment Loss} = $MSE$(IAF, stop-gradient(Q-IAF)) + $MSE$(OAF, stop-gradient(Q-OAF)) \;
    13. \textcolor{VQ}{VQ Loss} = $MSE$(Q-IAF, stop-gradient(IAF)) + $MSE$(Q-OAF, stop-gradient(OAF))\;
    14. Recon IF = Ins MLP\_Decoder(IAF) \tcp{shape=(1, $d$)} 
    15. Recon OF = Ins MLP\_Decoder(OAF) \tcp{shape=(1, $d$)} 
    16. \textcolor{RE}{Reconstruction Loss} = $MSE$(Recon IF, IF) + $MSE$(Recon OF, OF)\;
    17. \textcolor{matchingloss}{Matching Loss} = $Minimum_{i=1..k_1, j=1..k_2}$($MSE$($i_{th}$  IAF, $j_{th}$ OAF))
    }
    
\end{algorithm}
The full loss function is shown below:
\begin{multline}
    Loss = \lambda_1 \times Attribte\ Loss + \lambda_2 \times VQ\ Loss + \\
    \lambda_3 \times Commit\ Loss  + \lambda_4 \times Recon\ Loss + \lambda_5  \times Matching\ Loss
\end{multline}
where $\lambda_1$ is 2.0, $\lambda_2$ is 1.0, $\lambda_3$ is 0.25, $\lambda_4$ is 1.0, and $\lambda_5$ is 1.0. \textcolor{attributeloss}{Attribute Loss} provides a direct loss for directing the MLP Encoder to learn the projections of IF and OF to IAF and OAF~\footnote{The abbreviations in this section are inherited from the pseudo-code~\ref{pseudo_attribute}.}.
The next three items \textcolor{VQ}{VQ Loss}, \textcolor{CL}{Commitment Loss} and \textcolor{RE}{Reconstruction Loss} can be referred to VQ-VAE Loss~\citep{van2017neural}. Our motivation in using these three items is to provide indirect constraints that enable the IF and OF to be projected into the shared codebook's feature space.  To provide a direct alignment of the attribute features of a given instruction and objects that satisfy the given instruction, we design \textcolor{matchingloss}{Matching Loss}. An instruction and an object in the instruction's solution theoretically have at least one attribute that matches, \emph{i.e.}, an attribute of the object that necessarily satisfies part of this instruction; otherwise, the object should not be part of the instruction's solution. We consider the most similar pair of attribute features among the $k_1$ IAF and the $k_2$ OAF to be a match, and therefore need to reduce the error of this pair of attribute features.

In the end, only the instruction and object MLP Encoders and CLIP-Text-Encoder are used in navigation, and other part like codebook and MLP Decoder are not used.
Since this codebook is initialized by CLIP features and the ground-truth attribute features are also encoded by CLIP, the attribute feature space we get can actually be seen as a subspace of the CLIP semantic space. We hope that this design will improve the generalization of attribute features, since the CLIP semantic space shows good generalization and performance on many tasks~\cite{khandelwal2022simple,dorbala2022clip, gadre2023cows}.

\subsection{Coarse-to-fine Exploration Agent}

\label{c2f_exploration}
\begin{figure}[h]
  \centering
  \includegraphics[width=1\textwidth,trim=00 00 00 00,clip]{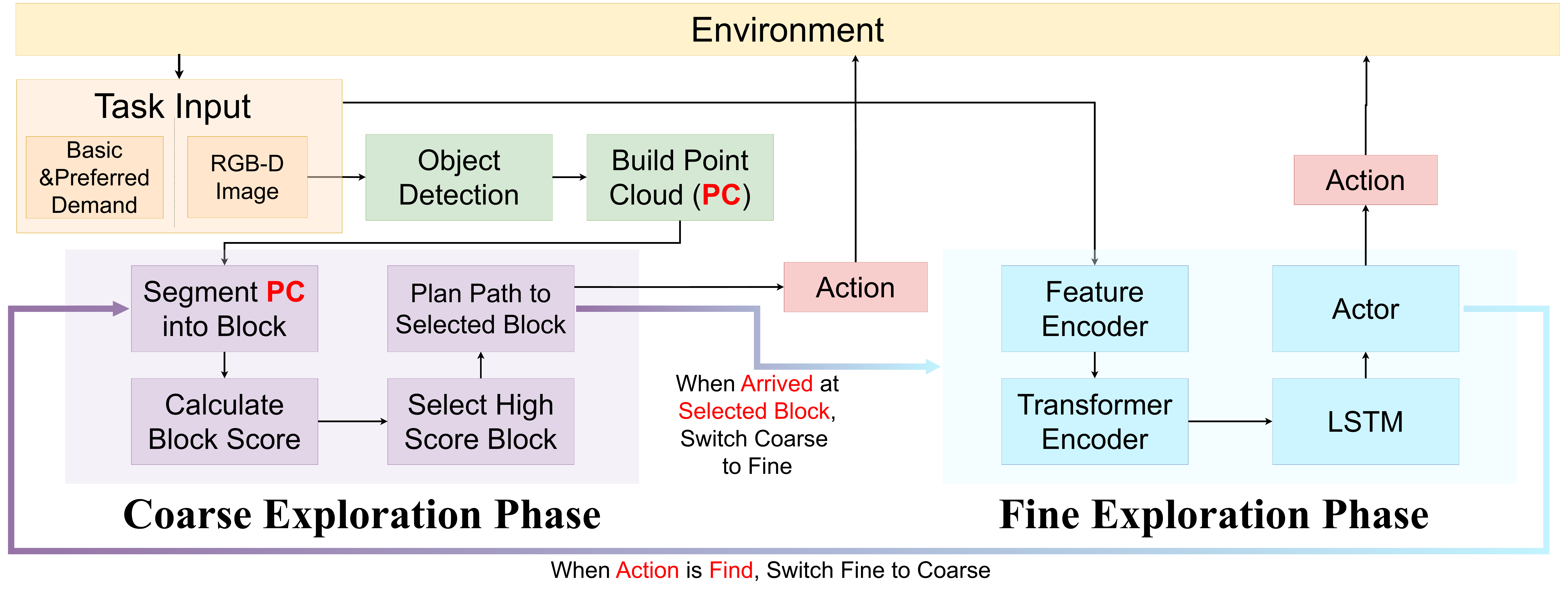}
  \caption{\textbf{Navigation Policy.} The agent continuously switches between a coarse exploration phase and a fine exploration phase until the $\mathrm{Find}$ count limit $n_{find}$ is reached or the total number of steps $n_{step}$ is reached. See Sec.~\ref{sec:coarse} and Sec.~\ref{sec:Fine} for details about the two phases. In each timestep, the GLEE model is used to identify and label objects in the RGB and project them to the point cloud.}
  \label{fig:switch}
\end{figure}

\begin{figure}[ht]
  \centering
  \includegraphics[width=0.95\textwidth,trim=00 00 00 00,clip]{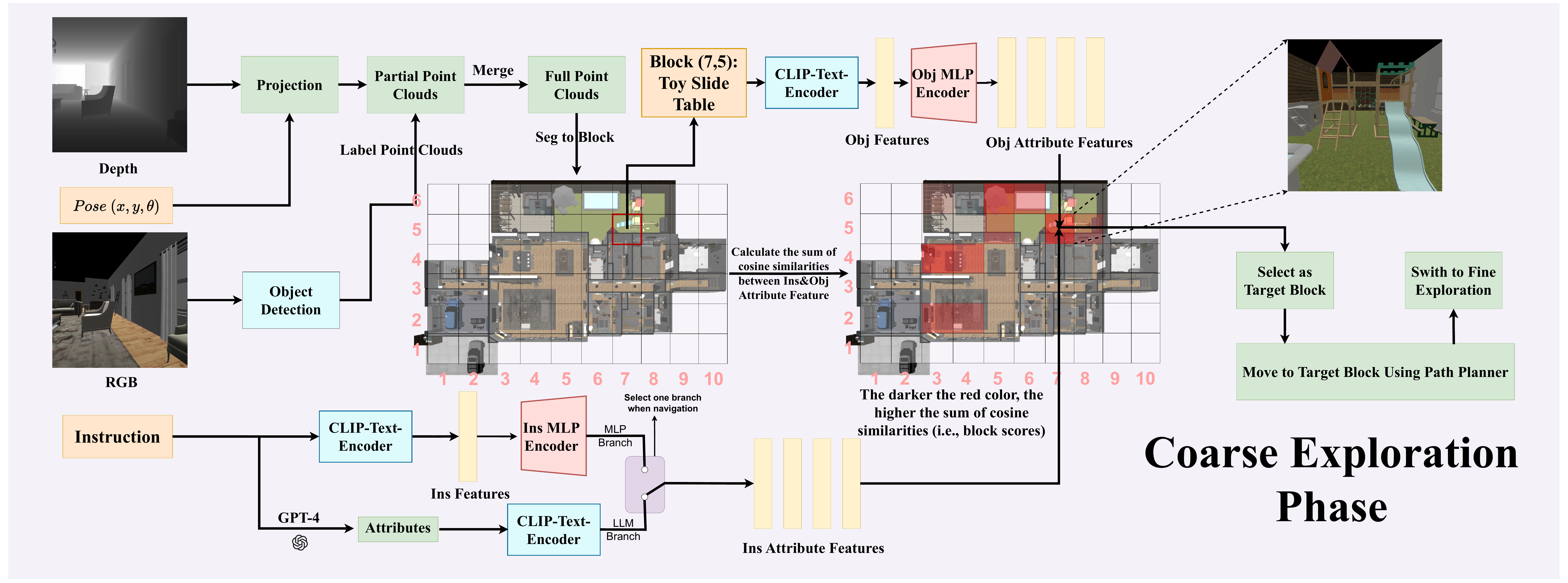}

  \caption{\textbf{Coarse Exploration.} This figure presents the process of building and labeling the point clouds, segmenting the blocks, and calculating the scores for each block.}
  \label{fig:coarse}
\end{figure}

\begin{figure}[ht]
  \centering
  \includegraphics[width=0.95\textwidth,trim=00 00 00 00,clip]{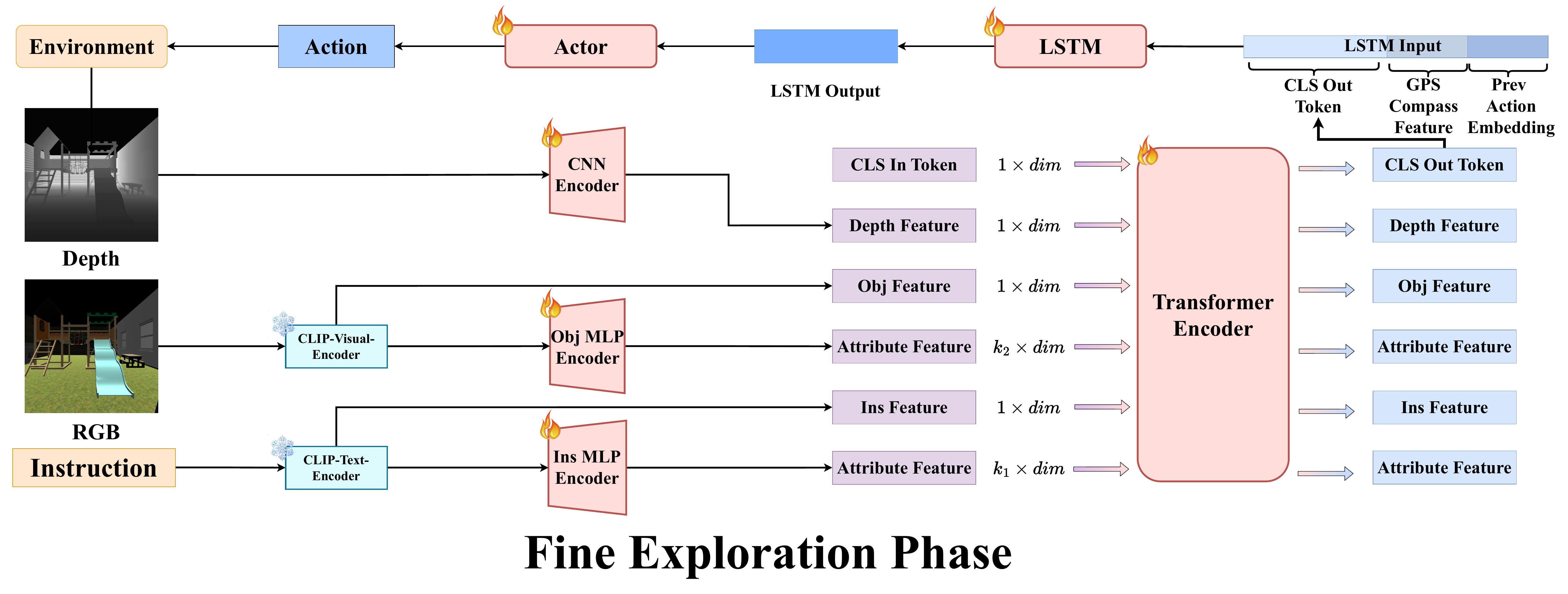}
  \caption{\textbf{Fine Exploration.} We employ imitation learning to train an end-to-end module in this phase. This module loads the Ins MLP Encoder and Obj MLP Encoder's parameters as initialization from attribute training, along with a Transformer Encoder to integrate features. The output feature corresponding to the CLS token is combined with GPS+Compass features and a previous action embedding and passed through an LSTM to generate actions by an actor.}
  \label{fig:fine}
\end{figure}

In this section, we describe how attribute features work in navigation. The agent switches between coarse and fine exploration phases back and forth until the number of $\mathrm{Find}$ executions reaches the upper limit $n_{find}$ or the number of steps reaches the upper limit $n_{step}$. Fig.~\ref{fig:switch} illustrates a general overview of the entire navigation policy and Fig.~\ref{fig:coarse} and Fig.~\ref{fig:fine} show the details of the coarse and fine exploration phase, respectively.
In both the coarse and fine exploration phase, we load the parameters of Ins MLP Encoder and Obj MLP Encoder from the attribute training for waypoint (\emph{i.e.}, block) selection and end-to-end fine exploration.
We argue that this coarse-to-fine design allows for the incorporation of prior knowledge from external foundation large models in the coarse exploration phase and task-relevant world-grounding exploration in the fine exploration phase.

\subsubsection{Coarse Exploration Phase}
\label{sec:coarse}
In this phase (see Fig.~\ref{fig:coarse}), the agent receives the RGB-D input, pose, and the demand instruction. We then use the camera parameters and depth map to compute partial point clouds of the current observation and merge them with the previously observed point clouds. We use an object detection model, GLEE, to detect objects in the RGB image and project them into the depth map, labeling the point clouds. We segment the point clouds into many rectangular blocks according to $x$ and $y$ coordinates, and each block is a $b\times b$ square, where $b$ is 2 in experiments. Thus, each detected object belongs to a block (according to the center of the object's point clouds). 
We use the instruction attribute features to query objects in each attribute block, and each block will then get a score. We use two different ways to generate instruction attribute features; see the LLM branch and MLP branch at the bottom in Fig.~\ref{fig:coarse}.
\textbf{For the LLM branch} in Fig.~\ref{fig:coarse}, we use GPT-4 to generate language-level basic and preferred attributes separately and use CLIP-Text-Encoder to obtain basic and preferred attribute features. These two attribute features can be used to calculate two scores, basic and preferred scores. Adjusting the weights of these two scores can control whether to prioritize the search for basic or preferred solutions.
The formula for calculating the score (\emph{i.e.}, the process of query) is as follows:
\begin{equation}
\label{score}
    s=\sum_{o \in block} (r_{p}\times \max_{i=1..k_1; j=1..k_2} f_{pref\_ins}^{i}*f_{o}^{j} + r_{b}\times \max_{i=1..k_1; j=1..k_2} f_{basic\_ins}^{i}*f_{o}^{j})
\end{equation}
Where $f_{o}^{j}$ denote $j_{th}$ attribute features of the object $o$, $f_{basic\_ins}^{i}$ denote $i_{th}$ basic attribute features of the instruction (so do $f_{pref\_ins}^{i}$), $*$ denotes cosine similarity, and $r_{b}$ and $r_{p}$ are adjustable weights for whether to find basic or preferred solutions.
If deployed in a real environment, these two weights can be freely adjusted by the user to apply to different situations. We conduct an ablation study on these two weights and discuss in detail how they affect the basic and preferred solutions in the experimental section.
\textbf{For the MLP branch}, we use the Ins MLP Encoder from attribute training to map the instruction features into $k_1$ attribute features, and we use a similar query process to compute scores.
MLP branch is a lightweight alternative that neither requires remote LLM nor consumes computational resources to run local LLMs. If deployed in a real environment, the agent is free to choose one of the branches according to the current LLM availability. We report the results of the two branches separately in the experimental section. 
Finally, we randomly choose a point in the highest-scoring and never-visited block as the waypoint. Then, the agent navigates to the point by a path-planning algorithm. 
Please see supplementary material for details about the coarse exploration module, the path-planning algorithm and blocks' score visualizations~\ref{supp:coarse_agent}.

\subsubsection{Fine Exploration Phase}
\label{sec:Fine}
In this phase, we train an end-to-end module using imitation learning similar to DDN. 
The alignment capabilities of CLIP in the visual and textual domains allow Obj MLP Encoder to still extract object attribute features through CLIP-Visual-Encoder's object features. The self-attention mechanism in the Transformer Encoder learns the association between the attribute features of the objects in the current field of view and the attribute features of the instructions, which can implicitly determine whether these objects need to be reported. Following BERT~\citep{devlin2018bert} and ViT~\citep{dosovitskiy2020vit}, we add a CLS token as the output of the feature fusion.
For more training and hyperparameter details, see the supplementary material~\ref{supp:fine_agent}.

\section{Experiment}
\label{exp}
\subsection{Experimental Settings}
We use habitat-sim and habitat-lab as our simulator and HSSD as our scene dataset. We randomly select 30 tasks as testing tasks and the remaining 270 tasks as training tasks(\emph{i.e.}, unseen task and seen task in Tab.~\ref{tab:main}, respectively). These training tasks are used to collect trajectories to train the fine exploration module, VTN and ZSON.
HSSD splits the scenes into val scenes and train scenes (\emph{i.e.}, unseen scenes and seen scenes in Tab.~\ref{tab:main}, respectively). In all experimental settings, the judgment distance of the found list $d_{find}$ is one meter, and a maximum number of $\mathrm{Find}$ $n_{find}$ is five times, and the maximum step number $n_{step}$ is 300.
A single RTX 4090 is enough to run the experiments. More experimental settings are available in the supplementary material~\ref{supp:exp}.

\subsection{Baselines}

\textbf{Random} is a method for randomly selecting actions. \textbf{VTN}~\citep{du2021vtnet} is an end-to-end closed-vocabulary ObjectGoal Navigation method. \textbf{ZSON}~\citep{majumdar2022zson} is an end-to-end open-vocabulary ObjectGoal Navigation method. \textbf{MOPA+LLM}~\citep{raychaudhuri2024mopa} is a modular-based Multi-object Navigation method using LLM to select target objects. \textbf{FBE+LLM}~\citep{yamauchi1997frontier} is a modular-based Multi-object Navigation method that uses frontier-based exploration (FBE) and LLM to find objects. \textbf{DDN}~\citep{wang2024find} is an end-to-end Single-object Demand-driven Navigation method.
 For more details about baselines in training and method modifications, please see supplementary material~\ref{supp:baseline}. 

\subsection{Baseline Comparison}
The experimental results are in Tab.~\ref{tab:main}. Random reflects the difficulty of MO-DDN. The two end-to-end ON methods, VTN and ZSON, perform poorly, only slightly higher than Random.  In comparison, though DDN uses attribute features designed for single object-instruction pairs, DDN's attribute features still provide some prior knowledge, outperforming VTN and ZSON. FBE+LLM and MOPA+LLM use different exploration strategies, but both use LLM to select waypoints. Both of them acquire the ground truth semantic label in the current RGB image when asking LLM whether or not to perform $\mathrm{Find}$\footnote{when other actions are performed, the ground truth semantic label will not be provided, and only the GLEE~\citep{wu2023general} detection will be provided.}, and they outperform the three end-to-end methods.
MOPA+LLM obtains suboptimal results in many cases, better than FBE+LLM, which may stem from the fact~\citep{raychaudhuri2024mopa} that its exploration strategy is superior to FBE, allowing more objects to be detected.

Our method outperforms the baseline in the vast majority of settings. Attribute similarity scores are a good substitute for LLM for waypoint selection. Compared to MOPA+LLM, which requires 5.15 requests to the LLM per episode on average, C2FAgent (LLM branch) only needs to request the LLM to parse instructions into attributes once at the beginning and achieves better performance. Moreover, C2FAgent (MLP branch) does not need LLM parsing and gets results comparable to MOPA+LLM. We also find that when no object meets the demand in the current detected object list, each block has a low score, approximating a randomly selected block to explore. In contrast, when there exists an object that meets the demand, the corresponding block has a significantly higher score than the other blocks. This indicates that the attribute model can balance exploration and exploitation on its own.

\begin{table}[]
\renewcommand{\arraystretch}{1.05}
\centering
\label{tab:main}
\caption{\textbf{Baseline Comparison.} Values in parentheses represent standard deviations. * represents the usage of ground truth semantic labels in the RGB image. The bold fonts represent optimal values.}
\resizebox{0.9\textwidth}{!}{
\begin{tabular}{c|cccccccc}
\toprule
\multirow{3}{*}{Method}                      & \multicolumn{8}{c}{Seen Scene}                                                                 \\ \cline{2-9} 
                                             & \multicolumn{4}{c|}{Seen Task}                           & \multicolumn{4}{c}{Unseen Task}     \\ \cline{2-9} 
                                             & $SR_b$ & $SR_p$ & $SPL_b$ & \multicolumn{1}{c|}{$SPL_p$} & $SR_b$ & $SR_p$ & $SPL_b$ & $SPL_p$ \\ \hline
Random                                       &  4.36 (0.36)  & 3.30 (0.07)  &  4.10 (0.45)  & \multicolumn{1}{c|}{2.92 (0.04)}        &   3.47 (0.27)     &  2.40 (0.22)      &    3.14 (0.26)     &    2.17 (0.25)     \\ \hline
VTN                                          &  7.50 (3.01) &  4.79 (3.02) &  4.19 (1.16)  & \multicolumn{1}{c|}{2.45 (1.51)}        &  5.17 (1.53)  &  4.06 (1.15)  & 3.38 (0.59) &  3.01 (0.99)  \\ \hline
ZSON                                         &  6.75 (2.61)  & 3.79 (0.82)  &  4.27 (1.95)  & \multicolumn{1}{c|}{2.41 (0.82)}        &  4.33 (1.26)      &    3.08 (0.75)    &   3.02 (1.22)   &   2.18 (0.65)      \\ \hline
DDN                                          & 8.10 (2.20)  &  6.62 (0.94) &  5.95 (2.09) & \multicolumn{1}{c|}{4.73 (1.20)}        &  6.70 (2.41)  & 4.52 (0.85)   &    4.67 (1.52) &  3.14 (0.96) \\ \hline
MOPA+LLM*                                      &  18.61 (3.75)   &  13.25 (1.99) &  3.00 (1.26)  & \multicolumn{1}{c|}{2.18 (0.58)}        &  14.67 (1.44) & \textbf{10.22 (1.28)}  &   2.03 (0.13)  &   1.55 (0.19) \\ \hline
FBE+LLM*                                      &   13.25 (0.35) & 9.95 (2.3)  &  3.9 (0.71) & \multicolumn{1}{c|}{3.24 (0.96)}        &   10.00 (1.80)  &  8.03 (2.23) & 3.37 (0.54)   & 2.99 (0.73)  \\ \hline
C2FAgent (MLP branch)                                      &  18.58 (2.78)  & 12.45 (2.30) &   6.99 (1.72)  & \multicolumn{1}{c|}{5.53 (1.05)}        &  13.91 (2.91) &  8.44 (0.87) &  4.45 (0.77)  &  3.43 (0.41) \\ \hline
C2FAgent (LLM branch)                                         &  \textbf{23.82 (3.89)} & \textbf{14.28 (3.72)} &  \textbf{7.94 (1.30)} & \multicolumn{1}{c|}{\textbf{5.80 (1.00)}}        &   \textbf{15.93 (1.68)} & 8.65 (0.71)  &   \textbf{6.21 (1.81)}  & \textbf{4.60 (1.40)}  \\ \hline
\midrule
\multicolumn{1}{l|}{\multirow{3}{*}{Method}} & \multicolumn{8}{c}{Unseen Scene}                                                               \\ \cline{2-9} 
\multicolumn{1}{l|}{}                        & \multicolumn{4}{c|}{Seen Task}                           & \multicolumn{4}{c}{Unseen Task}     \\ \cline{2-9} 
\multicolumn{1}{l|}{}                        & $SR_b$ & $SR_p$ & $SPL_b$ & \multicolumn{1}{c|}{$SPL_p$} & $SR_b$ & $SR_p$ & $SPL_b$ & $SPL_p$ \\ \hline
Random                                       &   6.60 (0.67)     &  4.47 (0.47)      &   6.02 (0.71)  & \multicolumn{1}{c|}{4.07 (0.44)}       &    4.63 (0.52)  &    3.46 (0.40)       &   4.18 (0.50)      &     3.18 (0.42)    \\ \hline
VTN                                          &   8.61 (4.25)  & 6.12 (3.84) &   4.76 (1.96)  & \multicolumn{1}{c|}{3.64 (1.86)}        &  6.83 (1.04)      &  3.47 (1.55)  &   3.48 (0.15)  &  1.97 (0.56) \\ \hline
ZSON                                         &  6.89 (1.50) &  5.06 (1.71) &   4.02 (1.19) & \multicolumn{1}{c|}{3.46 (1.92)}        &  5.33 (0.76) &  2.94 (0.51)  &  2.70 (1.02)  &   1.91 (0.57) \\ \hline
DDN                                          &  10.00 (2.50) & 5.93 (0.86) & \textbf{7.89 (1.93)}  & \multicolumn{1}{c|}{4.86 (0.52)}        &  7.10 (2.95) & 4.70 (1.81)    &  5.37 (2.46)  &  3.48 (1.45)  \\ \hline
MOPA+LLM*                                     &  17.83 (9.41)   &  14.17 (1.93) &  3.06 (0.53)  & \multicolumn{1}{c|}{2.96 (0.31)}        &  10.33 (2.08) & 6.72 (1.92)  &   1.19 (0.23)  &   0.78 (0.24) \\ \hline
FBE+LLM*                                      &   14.00 (4.47) & 8.03 (2.23)  &  3.13 (0.66) & \multicolumn{1}{c|}{2.99 (0.73)}        &   9.67 (1.53)  &  6.86 (1.04) & 1.78 (0.58)  & 2.12 (0.68)  \\ \hline
C2FAgent (MLP branch)                                         &  17.97 (2.87)  & 12.35 (2.49) &   5.49 (1.23)  & \multicolumn{1}{c|}{3.77 (0.70)}        &  12.17 (3.71) &  6.93 (1.63) &  3.51 (1.05)  &  2.46 (0.59) \\ \hline
C2FAgent (LLM branch)                                         &  \textbf{23.06 (1.58)} & \textbf{15.98 (0.87)}  &   6.52 (0.91)  & \multicolumn{1}{c|}{\textbf{5.11 (0.51)}}        &  \textbf{15.88 (2.63)}  & \textbf{9.48 (1.30)}  &  \textbf{6.03 (0.77)}  &   \textbf{4.50 (0.48)}      \\ \hline
\bottomrule
\end{tabular}}
\end{table}
\subsection{Ablation Study}

\begin{table}[h]
\centering 
\label{abla:result}
\begin{minipage}{0.47\textwidth}
\label{abla: coarse_Q1}
\caption{Ablation on Coarse Exploration (Q1)}
\centering 
\resizebox{0.94\textwidth}{!}{ 
\begin{tabular}{l|llll}
\toprule

Method             & $SR_b$ & $SR_p$ & $SPL_b$ & $SPL_p$ \\ \hline
\midrule
Coarse+Fine (Ours) &  \textbf{23.82 (3.89)} & \textbf{14.28 (3.72)} &  \textbf{7.94 (1.30)} &\textbf{5.80 (1.00)} \\ \hline
FBE+Fine        & 14.56 (5.57) &  11.36 (5.45)  &  4.94 (0.84)  & 3.92 (0.95)   \\ \hline
LLM+Fine         & 13.69 (4.91) & 9.47 (2.57)  &  5.12 (0.72)  &  3.32 (0.07) \\ \hline
CLIP+Fine         & 12.51 (3.85) & 7.65 (2.77)  &  4.52 (0.87)  &  3.05 (0.43) \\ \hline
\bottomrule
\end{tabular}}
\end{minipage}
\hfill 
\begin{minipage}{0.47\textwidth}
\label{abla: fine_Q2}
\caption{Ablation on Fine Exploration (Q2)}
\centering 
\resizebox{0.94\textwidth}{!}{ 
\begin{tabular}{l|llll}
\toprule

Method             & $SR_b$ & $SR_p$ & $SPL_b$ & $SPL_p$ \\ \hline
\midrule
Coarse+Fine (Ours)&  \textbf{23.82 (3.89)} & \textbf{14.28 (3.72)} &  \textbf{7.94 (1.30)} &\textbf{5.80 (1.00)} \\ \hline
Coarse+ZSON        & 8.74 (5.07)  &  4.21 (2.0) & 6.72 (4.29)  & 3.45 (2.0) \\ \hline
Coarse+VTN         & 16.89 (4.57) & 10.87 (4.73) & 5.36 (1.11)  & 3.83 (1.32)  \\ \hline
Coarse+Random      &  5.82 (1.22)   &  4.63 (0.93)   &  4.04 (2.52)    & 3.27 (2.12)   \\ \hline
\bottomrule
\end{tabular}}
\end{minipage}%
\end{table}
\begin{table}[ht]
\centering 

\begin{minipage}{0.47\textwidth}
\label{abla: attri_Q3}
\caption{Ablation on Attribute Training (Q3)}
\centering 
\resizebox{0.94\textwidth}{!}{ 
\begin{tabular}{l|llll}
\toprule

Method             & $SR_b$ & $SR_p$ & $SPL_b$ & $SPL_p$ \\ \hline
\midrule
Ours           &  \textbf{23.82 (3.89)} & \textbf{14.28 (3.72)} &  \textbf{7.94 (1.30)} &\textbf{5.80 (1.00)} \\ \hline
Ours w/o VQ-VAE        &  18.34 (1.33) &  11.42 (0.23)   &  5.63 (0.72)   &  4.72 (0.63)   \\ \hline
Ours w/o codebook init         & 17.36 (2.32) &  12.02 (2.54) &  5.73 (1.14) &  4.56 (1.18)  \\ \hline
\bottomrule
\end{tabular}}
\end{minipage}
\hfill 
\begin{minipage}{0.47\textwidth}
\label{abla: rate_Q4}
\caption{Ablation on Score Weights (Q4)}
\centering 
\resizebox{0.94\textwidth}{!}{ 
\begin{tabular}{l|llll}
\toprule

Method (C2FAgent)             & $SR_b$ & $SR_p$ & $SPL_b$ & $SPL_p$ \\ \hline
\midrule
$r_b=1, r_p=2$ &    19.89 (2.00)    & \textbf{15.05 (1.58)}  &  7.91 (1.32) &  \textbf{6.20 (0.64)} \\ \hline
$r_b=1, r_p=1$    &  23.82 (3.89) & 14.28 (3.72) &  7.94 (1.30) & 5.80 (1.00) \\ \hline
$r_b=1, r_p=0$         &  \textbf{25.10 (2.06)} &  12.43 (1.04)  &   \textbf{9.26 (2.07)}      &   5.34 (1.09)  \\ \hline
\bottomrule
\end{tabular}}
\end{minipage}%
\end{table}

In this section, we would like to discuss the following four questions:
\begin{itemize}
\item \textbf{Q1}: Is selecting waypoints by attribute feature similarity scores better than FBE, LLM and CLIP features' similarity scores? 
\item \textbf{Q2}: Do attribute features also work in the end-to-end fine exploration modules? How about replacing the fine exploration module with VTN and ZSON?
\item \textbf{Q3}: Do VQ-VAE losses and codebook initialization contribute to experimental results?
\item \textbf{Q4}: Can adjusting the weights of basic and preferred scores affect agent behavior?
\end{itemize}

In the ablation study, we report the results in the seen tasks and seen scenes. Ours refers to C2FAgent (LLM branch). The experimental results for all four questions are in Tab.~\ref{abla:result}.
For specific experimental setups, please see supplementary materials~\ref{supp:abla}.

\textbf{For Q1}, the experimental results demonstrate that attribute-based coarse exploration outperforms rule-based FBE, commonsense-based LLM and CLIP-based exploration. 
\textbf{For Q2}, the experimental results show that the fine exploration module utilizes the prior in the attribute model well, outperforming the VTN, which has larger model parameters, and the ZSON, which has been pre-trained on 36M total episodes and fine-tuned on the same trajectory dataset with Ours.
In addition, we note that Coarse+VTN exceeds VTN, suggesting that the coarse exploration module can steer the agent to the region where it is more likely to find objects that satisfy the demand.
\textbf{For Q3}, we find that the performance decreases after removing the VQ-VAE Loss or codebook initialization. Since the attribute model itself is trained on language-grounding tasks and thus agnostic to the task being evaluated, we can argue that the VQ-VAE Loss and the initialization contribute to the generalizability of attribute features.
\textbf{For Q4}, adjusting the weights of two scores can indeed affect the agent's behavior. By tuning up $r_p$, $SR_p$ and $SPL_p$ increase, while $SR_b$ and $SPL_b$ decrease; and vice versa. This characteristic allows the user to freely decide whether to prioritize the search for the basic or preferred solution in the current situation. For example, when a user who likes Coke is \textbf{very} thirsty, he can increase $r_b$ to allow the agent to satisfy the basic demand with a higher success rate, while in the general case, $r_p$ can be increased to allow the agent to try to satisfy the preferred demand. 

\section{Conclusion and Discussion}
\label{conclusion}
In this paper, we propose a new benchmark, MO-DDN, which can be regarded as a multi-object version of DDN. Moreover, MO-DDN can be considered a crucial preliminary step in task planning. 
We also propose a coarse-to-fine attribute-based exploration agent, extending the concept of ``attribute'' to a multi-object setting. The agent uses attribute features in different exploration phases. The experimental results show that our method outperforms the baselines, and the attribute features improve exploration efficiency.
The ablation study also demonstrated the effectiveness of our method. 

\paragraph{Limitations and Broader Societal Impacts } In this paper, we assume that the number of attributes of both instructions and objects is fixed values (\emph{i.e.}, $k_1$ and $k_2$). Though this facilitates training and its experimental results outperform baselines, there are still some gaps with real life. Future work could consider more flexible attribute features. Please see ~\ref{supp:limitation} for more limitation discussion.
To the best of our knowledge, there are no observable adverse effects on society.

\begin{ack}
This work was supported by The National Youth Talent Support Program (8200800081),  National Natural Science Foundation of China (No. 62376006) and National Natural Science Foundation of China (No. 62136001).  
\end{ack}

\bibliographystyle{IEEEtran}
\bibliography{reference}

\newpage
\appendix

\section{Appendix / supplemental material}
\begin{itemize}
    \item Multi-object Demand-driven Navigation Task~\ref{supp:moddn}
        \begin{itemize}
            \item Task Dataset Generation~\ref{task_generation}
            \item Task Metrics~\ref{srspl}
            \item Task Dataset Statistics~\ref{task_statistics}
        \end{itemize}
    \item Attribute Feature Training~\ref{supp:attribute_feature_training}
    \begin{itemize}
        \item Training Data Preparation~\ref{lgtask}
        \item Attribute Training Details~\ref{supp:attribute_training_method}
    \end{itemize}
    \item Details about Coarse-to-Fine Exploration~\ref{supp:c2f}
    \begin{itemize}
        \item Design Details and Visualizations for Coarse Exploration Module~\ref{supp:coarse_agent}
        \item Training Details about Fine Exploration Module~\ref{supp:fine_agent}
    \end{itemize}
    \item Experiments~\ref{supp:exp}
    \begin{itemize}
        \item Details about Experimental Settings~\ref{supp:exp_setting}
        \item Details about Baselines~\ref{supp:baseline}
        \item Details about Ablation Study~\ref{supp:abla}
        \item Details about LLM’s Prompt in Experiments~\ref{prompt_in_exp}
    \end{itemize}
    \item More Limitations~\ref{supp:limitation}
\end{itemize}
\subsection{Multi-object Demand-driven Navigation Task}
\label{supp:moddn}
\subsubsection{Task Dataset Generation}
\label{task_generation}
We base our instruction generation process on the 466 object categories in the HSSD dataset ~\cite{khanna2023hssd}.
In short, we perform three steps while generating the task: LLM pre-generation, LLM revision, and manual review (we use gpt-4-0125-preview API in this paper). In the first step, we prompt the LLM with task definition, available object category list, task format, and generation guidelines to generate the raw tasks. In the second step, we prompt the LLM with previous raw tasks one by one. We let the LLM check whether the input raw task is in the task format, whether the objects involved are in the given object category list, and whether the demand instructions and the solutions correspond to each other, and let the LLM state the reason for the revision before providing the result of the revision. In the third step, we manually check the generated tasks, removing overly far-fetched tasks and adding more obvious solutions. 

\label{task_example}
Here are some examples. For each example, we only show part of the solution.

\fbox{\parbox{\textwidth}{
task instruction: I need to display my photography collection, preferably with good lighting.\\
basic solution: [picture frame, bookshelf]\\
preferred solution: [picture frame, bookshelf, table lamp], [picture frame, bookshelf, ceiling lamp]
}
}

\fbox{\parbox{\textwidth}{
task instruction: I need a sleeping arrangement for a guest staying over one night; I hope the bed is comfortable enough.\\
basic solution: [single bed, blanket, pillow], [sofa, blanket]\\
preferred solution: [king bed, blanket, pillow], [double bed, blanket, pillow]
}
}

\fbox{\parbox{\textwidth}{
task instruction: Train my small dog with water and treats, preferably it is for pet only.\\
basic solution: [mixing bowl], [pet bowl], [bowl]\\
preferred solution: [pet bowl]
}
}

\fbox{\parbox{\textwidth}{
task instruction: I need to work on a writing project but prefer a quiet and comfortable space.\\
basic solution: [desk, swivel chair, notebook], [desk, straight chair, laptop]\\
preferred solution: [desk, swivel chair, notebook, earphone, room divider]
}
}

We use the following prompts to generate the raw tasks:

\label{task_prompt}
\doublebox{\parbox{\textwidth}{System Prompt: You are an AI assistant that can understand human demands and imagine what human demands can be met with existing object categories.}}

\doublebox{\parbox{\textwidth}{Prompt:\\
\#\#\# Task Generation: Demand-Driven Navigation \#\#\#\\
**Objective:** Create a navigation task where an agent must locate an object within a specific category that satisfies given demands and preferences.\\
**Category:**\\
bed, sofa, cup, desktop....\\
**Task Requirements:**\\
- **Basic Demand:** Describe the fundamental requirement for the object.\\
- **Preference:** Detail any additional preferences that refine the object selection.\\

\#\#\#Demand-driven Navigation Task Template\#\#\#\\
\{\\
    "task\_instruction": \$basic\_demand\$, \$preference\$\\
    "basic\_demand\_instruction": "String"\\
    "preferred\_demand\_instruction": "String"\\
    "basic\_solution": [[object\_a, object\_c], [object\_b]]\\
     "preferred\_solution": [[object\_a, object\_c, object\_d], [object\_b, object\_f]]\\
\}\\
This should be in dict format.\\
basic\_solutio is a list whose elements are lists, and each element represents a solution that meets the basic\_demand\_instruction; each solution may consist of one or more objects.\\
preferred\_solution is a list whose elements are lists, and each element represents a solution that meets both the basic\_demand\_instruction and preference; each solution may consist of one or more objects.\\
"[object\_g]" represents that just object\_g can meet the demand.\\
"[object\_x, object\_y, object\_z]" represents that only the combination of object x, y, and z can meet the demand.\\

 \#\#\# Example  \#\#\#\\
\$Example Tasks\$\\

 \#\#\#Additional Guideline \#\#\#\\
**Avoid demands related to needing a specific place or location.**\\
**Focus on generating novel demands, possibly in the realm of entertainment, that don't require a specific place.**\\
**Finalize the task using the provided template, ensuring it is concise and formatted correctly.**\\
**Whenever possible, generate task\_instruction that requires multiple combinations of objects to be met, such as [object\_a, object\_c, object\_d] and [object\_b, object\_c, object\_e].**\\

\#\#\# Previously Generated Task Instructions: \#\#\#\\
\$Previous Task Instruction Example\$\\

\#\#\# Process \#\#\# \\
Determine "basic\_demand\_instruction" and "preferred\_demand\_instruction" based on the object category.\\
Sequentially consider how each object or combination thereof meets the demands.\\
Finalize your response in the provided dict format, ensuring logical consistency between "basic\_solution" and "preferred\_solution." \\
Make the response more concise and clear  and can be executed as "json.loads(task)".\\
Please sequentially generate ten tasks split by "=========", and make sure the task is totally different from the previously generated task instructions.\\\\
}}

We use the following prompts to revise the raw tasks:

\doublebox{\parbox{\textwidth}{System Prompt: You are an AI assistant that can help check whether objects meet the demands, equipped with common life knowledge. Your reply should be in JSON string format.}}

\doublebox{\parbox{\textwidth}{Prompt:\\
\#\#\# Task Verification and Modification \#\#\#\\
**Objective:** Evaluate the provided task for logical consistency and feasibility. Modify it as necessary to ensure all requirements are met accurately.\\
\#\#\#Object Category\#\#\#\\
bed, sofa, cup, desktop....\\

\#\#\#Demand-driven Navigation Task Template\#\#\#\\
\{\\
    "task\_instruction": \$basic\_demand\$, \$preference\$\\
    "basic\_demand\_instruction": "String"\\
    "preferred\_demand\_instruction": "String"\\
    "basic\_solution": [[object\_a, object\_c], [object\_b]]\\
     "preferred\_solution": [[object\_a, object\_c, object\_d], [object\_b, object\_f]]\\
\}\\
This should be in dict format.\\
basic\_solution is a list whose elements are lists, and each element represents a solution that meets the basic\_demand\_instruction; each solution may consist of one or more objects.\\
preferred\_solution is a list whose elements are lists, and each element represents a solution that meets both the basic\_demand\_instruction and preference; each solution may consist of one or more objects.\\
"[object\_g]" represents that just object\_g can meet the demand.\\
"[object\_x, object\_y, object\_z]" represents that only the combination of object x, y, and z can meet the demand.\\

\#\#\#To be modified Task in JSON String Format\#\#\#\\
\$Task Dict\$\\

**Instructions:**\\
 1. Verify Object Categories: Ensure all items in 'basic\_solution' and 'preferred\_solution' are from the provided Object Category. Replace or remove any items that do not belong.\\
2. Expand Object Lists: Identify additional items within the Object Category that can fulfill the 'basic\_demand\_instruction' and 'preferred\_demand\_instruction'. Add these items to 'basic\_solution' and 'preferred\_solution' as appropriate.\\
 3. Validate Combination Solutions: Assess if each combination of objects in 'basic\_solution' meets the 'basic\_demand\_instruction' and if removing any object from these combinations makes them invalid. Apply the same verification for 'preferred\_solution' concerning both 'basic\_demand\_instruction' and 'preferred\_demand\_instruction'.\\
4. Ensure Subset Relationship: Confirm 'preferred\_solution' is a subset of 'basic\_solution'. If not, integrate missing combinations from 'preferred\_solution' into 'basic\_solution'.\\

**Evaluation and Modification Steps:**\\
 - Begin by reviewing the JSON string of the current task for any logical inconsistencies or missing elements.\\
- Identify and explain any aspects that are not rational or feasible within the context of the given Object Category.\\
- Propose modifications, ensuring all suggested items are included in the Object Category and adhere to the Demand-driven Navigation Task Template.\\
**Final Instruction:**\\
- Present a brief critique of the task's initial setup, highlighting any irrational elements.
- Offer a detailed plan for rectification, including specific changes to 'basic\_solution' and 'preferred\_solution'. Ensure these modifications respect the original Object Category and meet the template requirements.\\
- Your response should conclude with a revised JSON string format of the task, reflecting all necessary adjustments for coherence and completeness.\\

}}

\subsubsection{Task Metrics}
\label{srspl}
\begin{equation}
    SR_{basic}=\frac{1}{N} \sum^N_{i=1} \max_{s_b \in So_b} \frac{\sum_{o \in FL} \mathds{1}_{o \in s_b}}{Len(s_b)}
\end{equation}
An example of calculating $SR_{basic}$ in an episode: when the agent finds object [a, b, c, d, e, f] and the solutions are object [a, b, c, x, y] or [d, e, m, n], the $SR_{basic}$ is $max(\frac{len(a,b,c)}{len(a,b,c,x,y)},\frac{len(d,e)}{len(d,e,m,n)})$, \emph{i.e.}, 0.6.

We calculate the SPL~\cite{anderson2018evaluation} using the following formula:
\begin{equation}
    SPL_{basic}=\frac{1}{N} \sum^N_{i=1} SR_{basic}^{i} \frac{l_i}{max(p_i,l_i)}
\end{equation}
where $SR_{basic}^{i}$ is the success rate of  $i_{th}$ episode, \emph{i.e.}, $SR_{basic}^{i}=\max_{s_b \in So_b} \frac{\sum_{o \in FL} \mathds{1}_{o \in s_b}}{Len(s_b)}$, $l_i$ is the length of the shortest path of an episode, and $p_i$ is the length of the path taken by the agent in an episode.

\subsubsection{Task Dataset Statistics}
\label{task_statistics}
\begin{table}[]
\centering
\caption{A Comparison Between MO-DDN and DDN. The slash symbol "/" distinguishes between basic and preferred data. The left side of the slash indicates basic, and the right side indicates preferred.}
\label{tab:statis}
\begin{tabular}{c|cc}
\toprule
\multicolumn{1}{l|}{}                 & \multicolumn{1}{l}{MO-DDN}         & \multicolumn{1}{l}{DDN} \\ \hline
\midrule
Preference                            & \multicolumn{1}{c|}{\checkmark}           &                            \\ \hline
Multi-object                          & \multicolumn{1}{c|}{\checkmark}           &                            \\ \hline
Number of Object Category             & \multicolumn{1}{c|}{358}        & 109                        \\ \hline
Average Instruction Length            & \multicolumn{1}{c|}{17.44}      & 7.5                        \\ \hline
Average Number of Solution            & \multicolumn{1}{c|}{17.51/50.4} & 2.3                        \\ \hline
Average Number of Object Per Solution & \multicolumn{1}{c|}{2.41/3.59}  & 1                          \\ \hline
\bottomrule
\end{tabular}
\end{table}
In MO-DDN, we generate 300 tasks, encompassing 358 object categories from the HSSD dataset. Notably, the average instruction length in MO-DDN is 17.44, significantly longer than the average length of 7.5 in DDN.
 On average, there are 17.51 basic solutions and 50.4 preferred solutions per task (note that preferred solutions also include basic demands so that the number will be higher than just basic solutions), much more than the average solution number of 2.3 in DDN. Each basic solution will contain 2.41 objects, and each preferred solution will contain 3.59 objects, while only one object is in the solution of DDN. In general, MO-DDN increases the complexity of instructions and the diversity of solutions. This makes MO-DDN more relevant to real-life environments and more difficult. We create a table to compare the differences between DDN and MO-DDN shown in Tab.~\ref{tab:statis}.

 \subsection{Attribute Feature Training}
 \label{supp:attribute_feature_training}
\subsubsection{Training Data Preparation}
\label{lgtask}
Due to the restricted scope of object categories within the HSSD dataset, the tasks are limited to 358 object categories, a significantly smaller subset than those observed in real-world scenarios, which may cause overfitting on trained task and object categories. 
To address this limitation and enhance attribute feature generalization (\emph{i.e.}, we would like the attribute feature to be universal and not limited to the HSSD dataset), we introduce language-grounding tasks by prompting GPT-4 to generate tasks without constraints on object categories and learn attribute features over these language-grounding tasks. In these tasks, a broader array of objects and instructions is employed, thereby fostering improved diversity and generalization of attribute features. Then, for each language-grounding task, we ask GPT-4 to answer the questions ``What attributes are required to fulfill these demands?'' and ``What attributes are inherent to these objects?'' This process yields language-level attributes for instructions and objects, which is used for training attribute features.

We use the following prompts to generate language-grounding tasks:

\doublebox{\parbox{\textwidth}{System Prompt: You are a household task generator. Generate a list of ten daily household tasks or needs that a person might have. These should be tasks that are commonly encountered in a daily living environment.}}

\doublebox{\parbox{\textwidth}{Prompt:\\
\#\#\# Task Introduction \#\#\#\\
Generate a list of ten daily household tasks or needs that a person might have. These should be tasks that are commonly encountered in a daily living environment.\\

\#\#\# Task Example \#\#\#\\
\$Some Task Examples\$\\

\#\#\# Task Format \#\#\#\\
String: List[List[String]]\\
For example, task\_instruction: [["ObjectA", "ObjectB"], ["ObjectC", "ObjectD", "ObjectE"], ["ObjectF"], ["ObjectG", "ObjectH"]]\\

\#\#\#  Instructions \#\#\# \\
- Generate a list of 10 unique daily household tasks or needs and separate them with "=========".\\
- Generated ten tasks should be distinct from each other and Example.\\
- Each task should be clear, concise, and understandable for an agent to execute or recognize.\\
- Avoid repetitive tasks; ensure each one is distinct.\\
- Consider a variety of household needs, including cleaning, maintenance, personal care, and organizational tasks.\\
- For each task, generate ten different objects or objects' combinations (e.g., [ObjectA, ObjectB] is a combination, [ObjectC, ObjectD, ObjectE] is another combination, [ObjectF] is a single object, [ObjectG, ObjectH] is another combination) that can satisfy the demand or be used to complete the task.\\
- Follow the format of the example provided in the Task Format.\\
- Each Object should be enclosed in double quotes, for example, "ObjectA".\\
}}

Here are some language-grounding tasks:

\fbox{\parbox{\textwidth}{task instruction: I want to exercise at home, focusing on my abdominal muscles, preferably with minimal noise.\\
basic solution: [exercise bike],[yoga mat, dumbbell]\\
preferred solution:[yoga mat]
}}

\fbox{\parbox{\textwidth}{task instruction: I need a secure way to store small valuables at home, preferably in a method that's not immediately obvious to guests.\\
basic solution: [safe],[bookcase, book]\\
preferred solution:[bookcase, book]
}}

\fbox{\parbox{\textwidth}{task instruction: I need an efficient method to keep track of my daily hydration, preferably with a solution that's easy to transport.\\
basic solution: [water bottle],[thermos]\\
preferred solution:[water bottle]
}}

\fbox{\parbox{\textwidth}{task instruction: I want to relax in my garden with a chair that gently rocks, but I prefer it to be able to withstand the weather conditions.\\
basic solution: [rocking chair, gazebo], [camp chair, umbrella]\\
preferred solution:[rocking chair, gazebo]
}}

We can see that the language-grounding task uses some objects that do not appear in the HSSD dataset. We then generate several attributes for each task's instructions and objects. 

We use the following prompts to generate attributes for instructions:

\doublebox{\parbox{\textwidth}{System Prompt: You are a household attribute generator. Generate a list of attributes or functions that can be used to satisfy the given instructions.}}

\doublebox{\parbox{\textwidth}{Prompt:\\
 \#\#\# Attribute Introduction \#\#\#\\
Every object or item in the world has some attributes or functions that can be used to meet human demands. For example, water has the properties or functions of being drinkable, usable for washing clothes, capable of dissolving some solids, and suitable for bathing. \\
When a human need is satisfied by a certain object or item, it is essentially the attributes of that object that meet this demand. Therefore, we can deduce the required attributes or functions from a specific demand.\\

\#\#\# Attribute Format \#\#\#\\
    String: List[String]\\
    For example, "instruction": ["Attribute1", "Attribute2", "Attribute3", "Attribute4"]\\

\#\#\# Example \#\#\#\\
\$Some Examples\$\\

\#\#\# The Given Instruction \#\#\#\\
\$Instruction\$\\

\#\#\# Instructions \#\#\#\\
 for each demand, generate four attributes or functions that can be used to satisfy the given demand, separated by "=============". \\
 Instructions and Attribute should be separated by ": ".\\
 Each attribute or function should be clear, concise, and understandable for an agent to execute or recognize.\\
 Avoid using words that are too general or vague.\\
 The attribute or function can be a word or phrase.\\
 Avoid using synonyms or similar words that can be used interchangeably.\\
 Consider a variety of attributes, including personal, organizational, and functional attributes.\\
 Follow the Attribute Format provided above.\\
 Maintain the order of The Given Instruction same with the original.\\
}}
Here are instructions and their attributes.

\fbox{\parbox{\textwidth}{task instruction: I want to exercise at home without assembling bulky equipment, preferring compact and versatile fitness tools.\\
attributes: [resistance bands,
            adjustable dumbbells,
            foldable yoga mat,
            doorway pull-up bar]
}}

\fbox{\parbox{\textwidth}{task instruction: I want to easily access my jewelry collection and minimize clutter, preferably by displaying the pieces.\\
attributes:[
            hanging organizers,
            jewelry stands,
            "drawer dividers,
            wall-mounted display]
}}

\fbox{\parbox{\textwidth}{task instruction: I want to keep my cosmetics organized and easily accessible, preferably in a tidy arrangement.\\
attributes:[
            compartmentalized storage,
            transparent containers,
            countertop design,
            cosmetic organizers]
}}

\fbox{\parbox{\textwidth}{task instruction: I want a simple way to track grocery needs, preferably one that the whole family can access.\\
attributes:[
            shared list,
            real-time updating,
            accessible by multiple users,
            simple interface]
}}

We use the following prompts to generate attributes for objects:

\doublebox{\parbox{\textwidth}{System Prompt: You are a household attribute generator. Generate a list of attributes or functions that can be used to satisfy the given demand.}}

\doublebox{\parbox{\textwidth}{Prompt:\\
\#\#\# Attribute Introduction \#\#\#\\
Every object or item in the world has some attributes or functions that can be used to meet human demands. For example, water has the properties or functions of being drinkable, usable for washing clothes, capable of dissolving some solids, and suitable for bathing. \\
When a human need is satisfied by a certain object or item, it is essentially the attributes of that object that meet this demand. We can obtain the attributes and functions of an object from common sense or experience.\\

\#\#\# Attribute Format \#\#\#\\
    String: List[String]\\
    For example, Object1: ["Attribute1", "Attribute2", "Attribute3", "Attribute4"]\\

\#\#\# Example Object \#\#\#\\
water: ["drinkable", "usable for washing clothes", "capable of dissolving some solids", "suitable for bathing"]=============\\
apple: ["sweet", "edible", "vitamin C", "vitamin A"]=============\\
book: ["readable", "hold information", "hold data," "hold text"]=============\\
shirt: ["comfortable", "cloth", "wearable", "can be worn"]=============\\

\#\#\# Example Attribute \#\#\#\\
\$ Example Attributes\$\\

\#\#\# The Given Objects \#\#\#\\
\$ Object Category\$\\

\#\#\# Guidance \#\#\#\\
 For each object, generate four attributes or functions that can be used to satisfy a demand, separated by "=============". \\
 Object and Attribute should be separated by ": ".\\
 Each attribute or function should be clear, concise, and understandable for an agent to execute or recognize.\\
 Avoid using words that are too general or vague.\\
 The attribute or functions can be a word or phrase.\\
 Avoid using synonyms or similar words that can be used interchangeably.\\
 Consider a variety of attributes, including personal, organizational, and functional attributes.\\
 Follow \#\#\# Attribute Format \#\#\# provided above. Follow the content format of \#\#\# Example Attribute \#\#\#.\\
 Maintain the Given object unchanged and only generate the attributes or functions.\\
}}

Here are objects and their attributes.

\fbox{\parbox{\textwidth}{object: long-handled duster\\
attributes:[
        extended reach cleaning,
        dust-attracting fibers,
        flexible head for tight spaces,
        washable and reusable]
}}

\fbox{\parbox{\textwidth}{object: rocking chair\\
attributes:[
       relaxation,
        comfortable seating,
        aesthetic addition to room,
        soothes babies]
}}

\fbox{\parbox{\textwidth}{object: adjustable table\\
attributes:[
       height customization,
        ergonomic design,
        collapsible for storage,
        multi-use surface]
}}

\fbox{\parbox{\textwidth}{object: dusting brush\\
attributes:[
       removes surface dust,
        gentle on delicate items,
        reaches tight spaces,
        ergonomic handle]
}}

\subsubsection{Attribute Training Details}
\label{supp:attribute_training_method}

The full loss function is shown below:

\begin{multline}
    Loss = \lambda_1 \times Attribte\ Loss + \lambda_2 \times Matching\ Loss + \\
    \lambda_3 \times VQ\ Loss + \lambda_4 \times Commit\ Loss + \lambda_5 \times Recon\ Loss 
\end{multline}
where $\lambda_1$ is 2.0, $\lambda_2$ is 1.0, $\lambda_3$ is 1.0, $\lambda_4$ is 0.25, and $\lambda_5$ is 1.0.

\subsection{Details about Coarse-to-Fine Exploration}
\label{supp:c2f}
\subsubsection{Design Details for Coarse Exploration Module}
\label{supp:coarse_agent}
For the LLM branch, GPT-4 generates the attributes of the instruction in natural language. For example, ``I need a comfortable place to read, preferably with natural light.'' has basic attributes of ``comfortable seating'', ``quiet environment'' and has preferred attributes of ``natural light''. Then, we use the CLIP-Text-Encoder to encode these language-level attributes into the instruction attribute features. 
For the MLP branch, we use CLIP-Text-Encoder to encode the instruction into the instruction features and then use Ins MLP Encoder to encode the instruction features into the instruction attribute features. 

For the path-planning algorithm, we use the habitat-sim built-in greedy planner. We ensure each step is on an already explored point clouds to avoid navigating through unexplored points. In this paper, for a fair comparison, baselines and ablations that involve building maps and traveling on them will all use habitat-sim's built-in greedy planner, including but not limited to MOPA+LLM and FBE+LLM.

\begin{figure}[ht]
  \centering
  \includegraphics[width=1\textwidth,trim=00 00 00 00,clip]{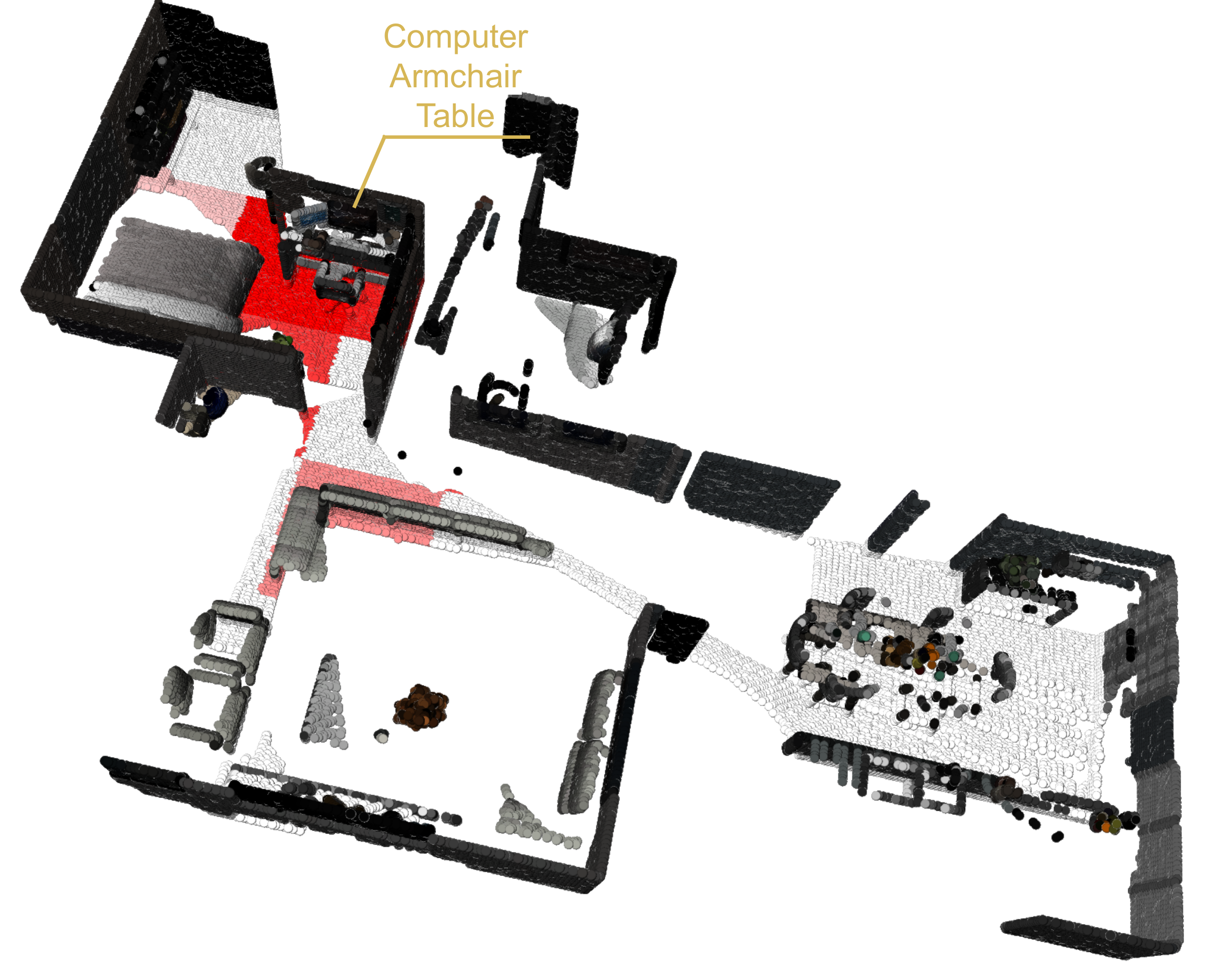}
  \caption{\textbf{Block Visualizations.}
  Instruction: I need to take quick notes during a meeting, preferably with a device that saves them digitally.
  Solution: Computer, Armchair, Table.
  }
  \label{fig:19}
\end{figure}

\begin{figure}[ht]
  \centering
  \includegraphics[width=1\textwidth,trim=00 00 00 00,clip]{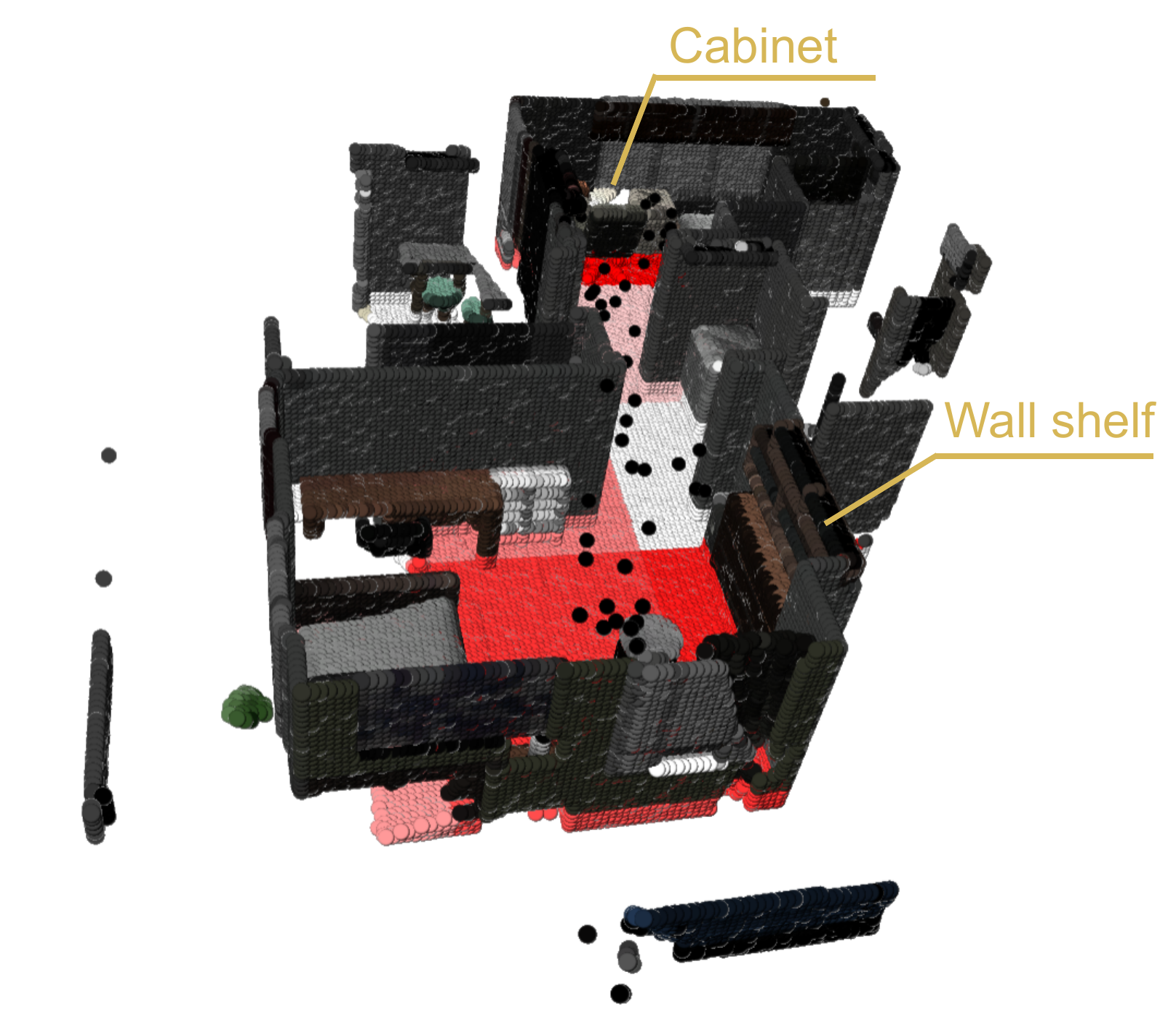}
  \caption{\textbf{Block Visualizations.}
  Instruction: I need to store a collection of fine china, preferably in a way that displays them elegantly.
  Solution: Wall shelf, Cabinet.
  }
  \label{fig:12}
\end{figure}

\begin{figure}[ht]
  \centering
  \includegraphics[width=1\textwidth,trim=00 00 00 00,clip]{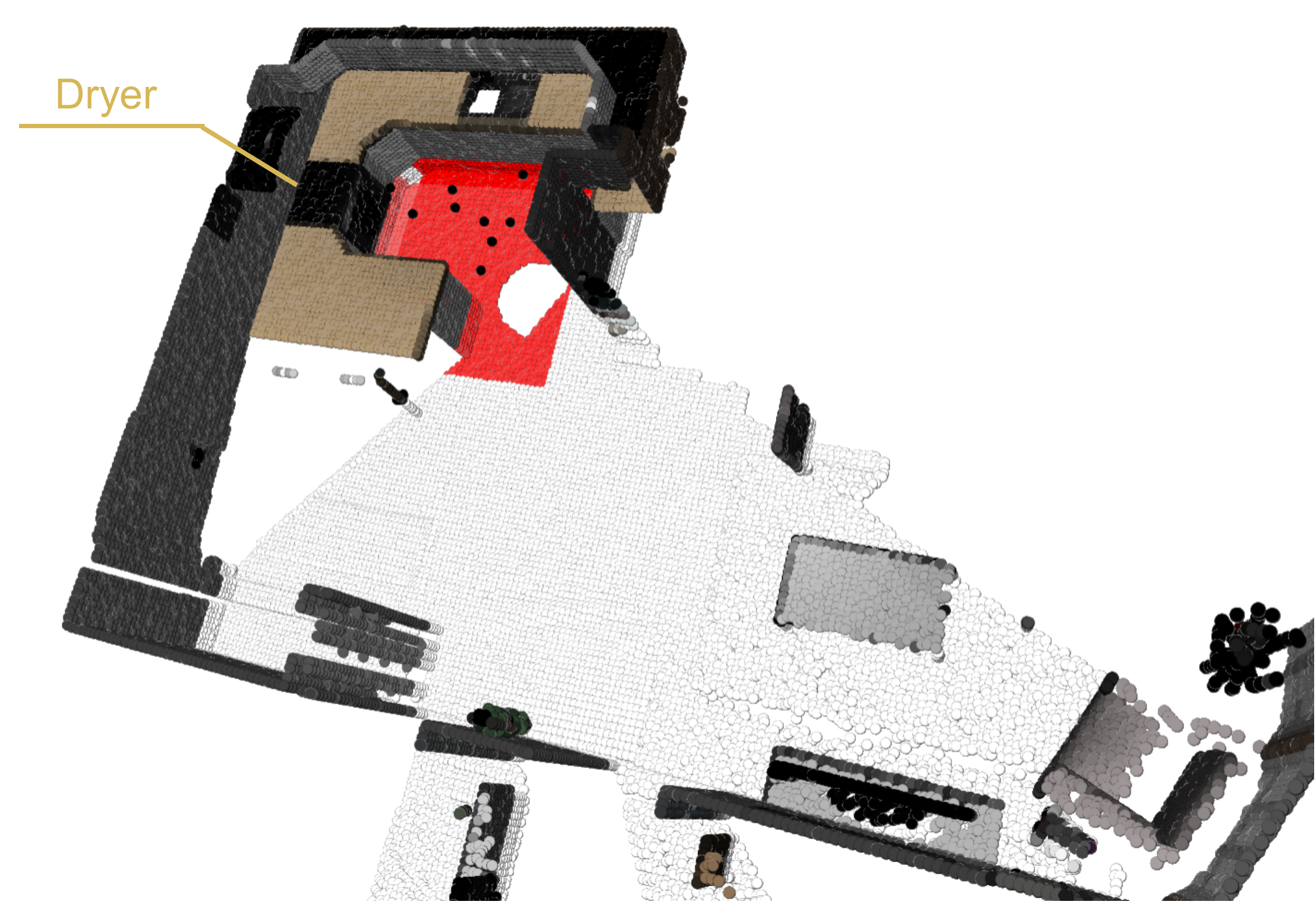}
  \caption{\textbf{Block Visualizations.}
  Instruction: I need to quickly dry a batch of laundry, but I prefer an fast and energy-efficient method.
  Solution: Dryer.
  }
  \label{fig:5}
\end{figure}

\begin{figure}[ht]
  \centering
  \includegraphics[width=1\textwidth,trim=00 00 00 00,clip]{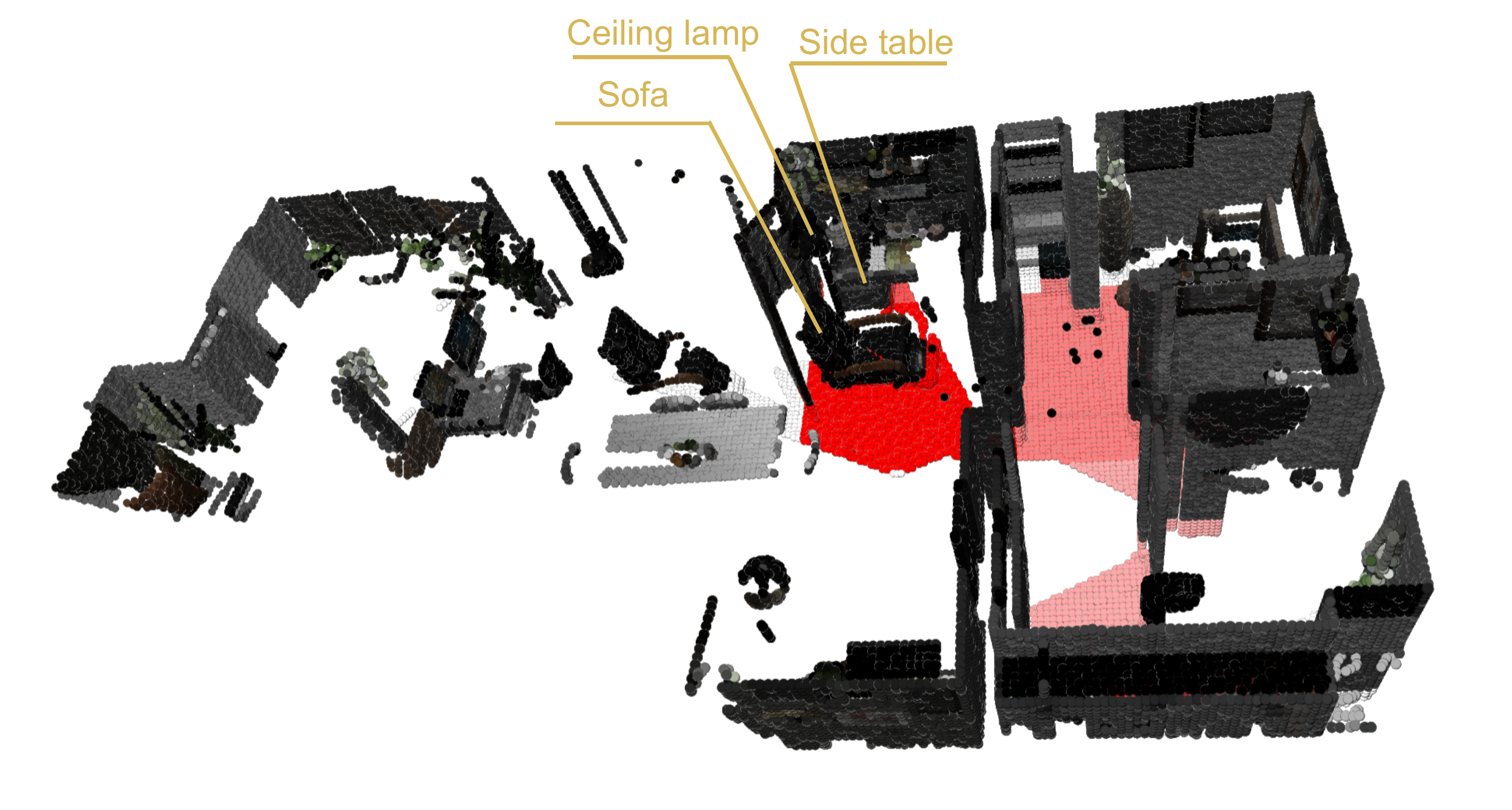}
  \caption{\textbf{Block Visualizations.}
  Instruction: I need to find a comfortable place to read for my study group, preferable with good lighting.
  Solution: Sofa, Side table, Ceiling lamp.
  }
  \label{fig:3}
\end{figure}

\begin{figure}[ht]
  \centering
  \includegraphics[width=1\textwidth,trim=00 00 00 00,clip]{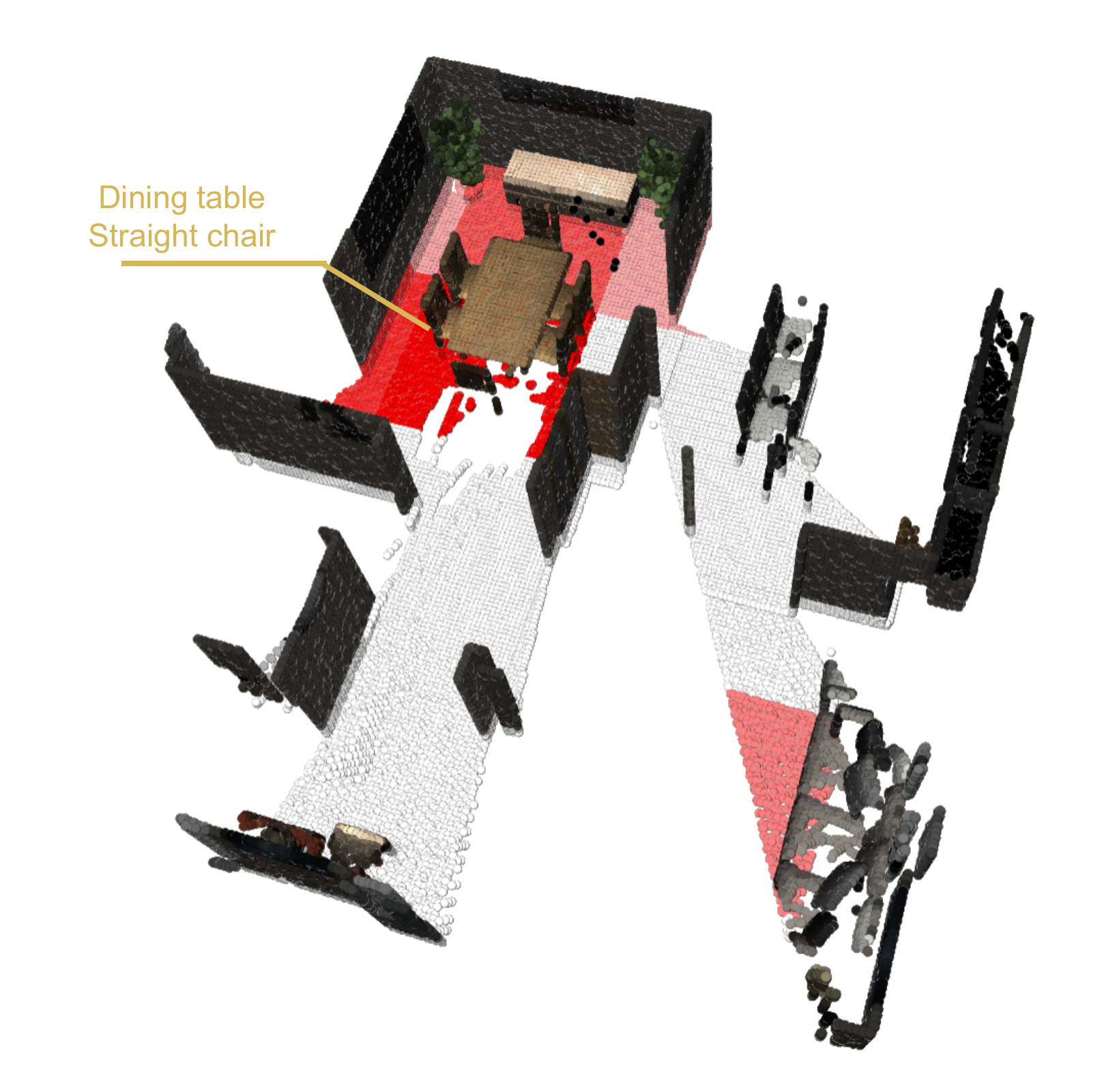}
  \caption{\textbf{Block Visualizations.}
  Instruction: I need to organize a small evening gathering but want to avoid any accidents with real candles.
  Solution: Dining table, Straight chair.
  }
  \label{fig:4}
\end{figure}

\begin{figure}[h]
    \centering
    \includegraphics[width=0.9\textwidth,trim=00 00 00 00,clip]{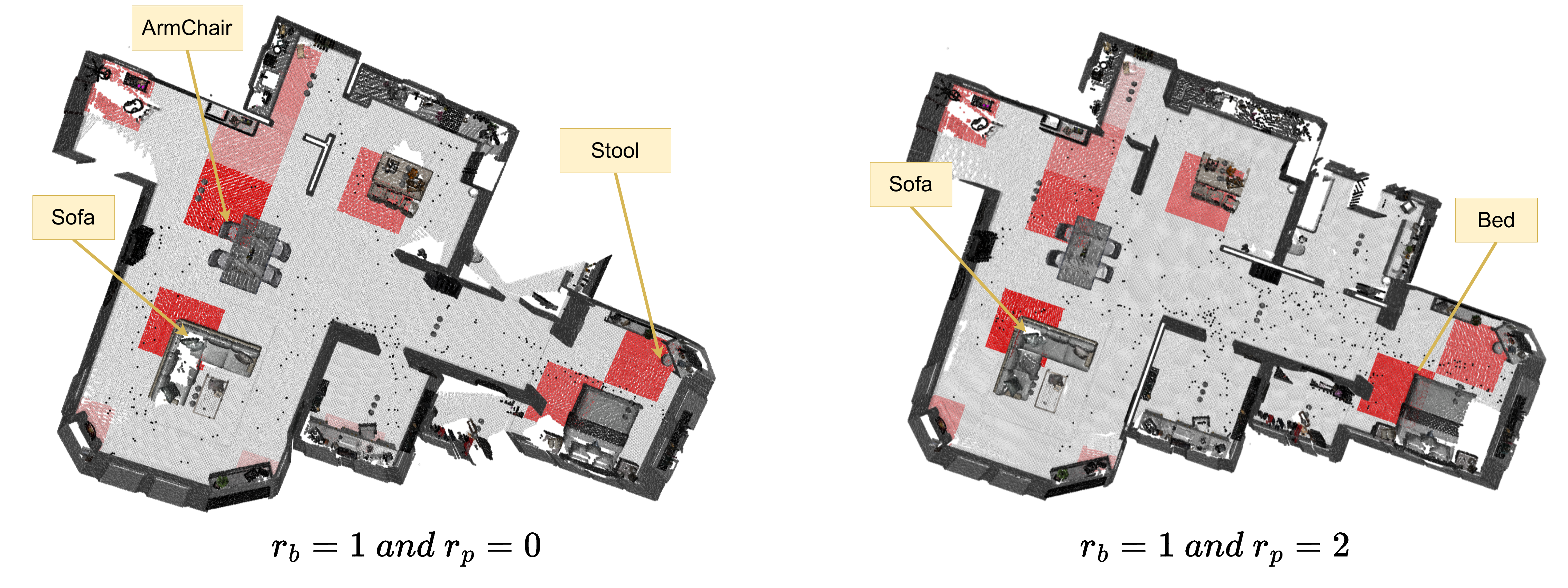} 
    \caption{\textbf{The Effect of $r_b$ and $r_p$}. This is an example used only to demonstrate scoring blocks. The darker the color in the figure, the higher the block score. When the task is ``Find a place to sit, prefering a comfortable and soft place'', lowering $r_p$ (left) ignores preferences, so ArmChair, Sofa and Stool can get a high score (dark red); raising $r_p$ (right) focuses on preferences, so Sofa and Bed can get a high score.}
    \label{fig:basic_vis}
    
\end{figure}

We provide some block score visualizations. A darker red color means a higher cosine similarity between the attribute features of the objects and the instruction. The darkness of the color is based on the rank of the block score among all blocks, not the actual score. Since scores are not calculated for blocks that have been visited before, only a single coarse exploration is included in the visualization. Note that the block with the highest score (\emph{i.e.}, the darkest color) is likely to contain objects that can satisfy the demands.
See Fig.~\ref{fig:19}, Fig.~\ref{fig:12}, Fig.~\ref{fig:5}, Fig.~\ref{fig:3}, Fig.~\ref{fig:4}. We also provide a visualization of block scores with different weights at Fig.~\ref{fig:basic_vis}.

\subsubsection{Training Details about Fine Exploration Module}
\label{supp:fine_agent}
We use the standard transformer encoder from the official PyTorch 1.13.1 implementation, where d\_model is 768, nhead is 8, num\_layers is 6, and other parameters remain default. 
The embedding dim of action is 64. The embedding dim of GPS+Compass is 32. The input dim of LSTM is 768+64+32, its hidden\_size is 1024, and its num\_layers is 2.
The depth model is a simple five-layer CNN model and a two-layer MLP model.

We collect about 50,000 trajectories to train the fine exploration module under seen tasks and seen scenes settings in Experiments Sec.~\ref{exp} using imitation learning. We collect trajectories by following steps: 1) randomly select a scene and a task, 2) initialize agent within two meters (\emph{i.e.}, block size in coarse exploration) of a target object, 3) use habitat-sim's built-in greedy planner to get the next step, 4) when the distance to the target object is less than 0.2 meters, turn left/right or look up/down according to the height and position of the object, 5) when the target object is in the field of view, execute $\mathrm{Find}$ and close this trajectory. The trajectories we collect here are relatively short and end with only one execution of $\mathrm{Find}$. The average length of these trajectories is 9.19.
This is because the fine exploration module only needs to take on exploration within a block, and therefore, the goal of learning only needs to be a trajectory within a block. Training with such short trajectories is very efficient and requires few computational resources (only a RTX 4090 graphics card is needed).
The standard deviation of the trajectory length is 5.62. The median trajectory length is 8. The plurality of trajectory lengths is 3. The maximum trajectory length is 51. The minimum trajectory length is 2.
We trained the model on a single RTX 4090 using imitation learning and cross-entropy loss, i.e., considering the action prediction as a classification task, consuming about 12h.

We keep switching between the two exploration phases until the number of $\mathrm{Find}$ reaches $n_{find}$, and then execute $\mathrm{Done}$. When $\mathrm{Find}$ is executed in the fine exploration phase, switch back to the coarse exploration phase to continue selecting the next waypoint.

\subsection{Experiments}
\label{supp:exp}
\subsubsection{Details about Experimental Settings}
\label{supp:exp_setting}
The effect of each action is as follows:

\begin{itemize}
    \item $\mathrm{MoveAhead}$: Move forward 0.25 meters.
    \item $\mathrm{RotateRight}$: Turn right 30 degrees.
    \item $\mathrm{RotateLeft}$: Turn left 30 degrees.
    \item $\mathrm{LookUp}$: Turn the camera up 30 degrees.
    \item $\mathrm{LookDown}$: Turn the camera down 30 degrees.
    \item $\mathrm{Find}$: Record objects with a distance less than $d_{find}$ in the current field of view to the found list.
    \item $\mathrm{Done}$: End the current episode and report the success rate and SPL.
\end{itemize}

Note that we want to focus the benchmark on navigation. When executing $\mathrm{Find}$, the simulator will automatically add objects to the found list that have distances less than $d_{find}$ in the current field of view, regardless of whether or not they are recognized by the object detection module GLEE. Such detection is done by the simulator, so the agent does not get the ground truth semantic label, except when $\mathrm{Find}$ determinations are made by MOPA+LLM and FBE+LLM. 

Our method and baselines can be trained on a single RTX 4090, which will take about one day for each method. 
 We report the mean and standard deviation for at least three seeds. We test the agent for 100 epochs in each seed, which takes about 8 hours.

\subsubsection{Details about Baselines}
\label{supp:baseline}

Since many baselines are policies in a single object navigation setting, we make some modifications to their action space. In their original policy, $\mathrm{Done}$ is modified to $\mathrm{Find}$, and $\mathrm{Done}$ is automatically output when the number of $\mathrm{Find}$ executions reaches the upper limit $n_{find}$.

For Random, we let the agent randomly select an action other than $\mathrm{Done}$ and execute it. When the number of $\mathrm{Find}$ reaches the maximum number of executions $n_{find}$ or the number of steps reaches the limit, execute $\mathrm{Done}$.

For VTN, it is trained in the original paper in two stages, imitation learning without LSTM and reinforcement learning with LSTM. We remove the reinforcement learning phase here because we find that in the MO-DDN task, using reinforcement learning to train VTN leads to performance degradation. We, therefore, chose to collect 20,000 trajectories and train the VTN using imitation learning with an LSTM module. We collect trajectories by following these steps: 1) randomly select a scene and a task, 2) initialize agent at a random position, 3) set a target object according to the solution, 4)  use habitat-sim's built-in greedy planner to get the next step, 5) when the distance to the target object is less than 0.2 meters, turn left/right or look up/down according to the height and position of the object, 6) when the target object is in the field of view, execute $\mathrm{Find}$, 7) go back to step 3 until all objects in the solution have been searched for, and then execute $\mathrm{Done}$. The average length of these trajectories is 49.15. We replace the VTN's original goal description with the CLIP features of the demand instruction.
During evaluating, the agent executes $\mathrm{Done}$ when the number of $\mathrm{Find}$ reaches the maximum number of executions $n_{find}$ or the number of steps reaches the limit.

For ZSON, we use imitation learning to fine-tune the official ZSON's trained weights using the trajectories collected in the VTN. We take the CLIP features of the demand instruction as ZSON's goal descriptions.
During evaluating, the agent executes $\mathrm{Done}$ when the number of $\mathrm{Find}$ reaches the maximum number of executions $n_{find}$ or the number of steps reaches the limit. 

For DDN, we train the agent using the trajectories collected in the VTN. For training the attribute model in DDN, we concatenate all objects in a solution at the language level and encode them with CLIP-Text-Encoder as object features. If two solutions satisfy the same demand instruction, then the object features corresponding to these two solutions are positive samples. During evaluating, the agent executes $\mathrm{Done}$ when the number of $\mathrm{Find}$ reaches the maximum number of executions $n_{find}$ or the number of steps reaches the limit.

For MOPA+LLM, we use map-building and object-detection models that are the same as our method, and we use LLM to make selections for currently recognized objects. When the LLM selects an object as a target, we use habitat-sim's built-in greedy planning to walk near it; if the LLM does not select any object, it chooses a random point on the already explored map. Compared to the original MOPA implementation, we replace the point goal agent with the built-in greedy planner of habitat-sim. When the agent reaches the target, we prompt the LLM to decide whether to perform the $\mathrm{Find}$ action. The prompts we use are in Sec.~\ref{prompt_in_exp}.

For FBE+LLM, we use the FBE as the exploration module and then let the LLM choose which objects that have been explored meet the demand and serve as targets for the habitat-sim's built-in greedy planner to walk around. If the LLM does not select an object, the agent continues to explore using the FBE method. When the agent reaches the target, we prompt the LLM to decide whether to perform the $\mathrm{Find}$ action. The prompts we use are in Sec.~\ref{prompt_in_exp}.

\subsubsection{Details about Ablation Study}
\label{supp:abla}
\paragraph{Ablation on Coarse Exploration} For FBE+Fine, we use FBE to select waypoints and then use the habitat-sim's built-in greedy planner to get to the waypoints. We use the fine exploration module to locate the target objects when arriving at the waypoints. For LLM+fine, we use LLM to select waypoints, and everything else is the same as FBE+Fine. The prompts we use are in Sec.~\ref{prompt_in_exp}. For CLIP+fine, we replace attribute features with CLIP features to compute cosine similarity as block scores.

\paragraph{Ablation on Fine Exploration} For Coarse+ZSON/VTN, we use the same trajectory dataset used to train the fine exploration module to train ZSON and VTN and then replace the fine exploration module with ZSON/VTN. For Coarse+Random, we replace the fine exploration module with randomly selecting an action except for $\mathrm{Done}$.

\paragraph{Ablation on Attribute Training} For Ours w/o VQ-VAE, we set the weight of VQ Loss, Commit Loss, and Recon Loss to zero and only maintain Attribute Loss and Matching Loss. For Ours w/o initialization, we do not load the clustering centers to codebook. 

\paragraph{Ablation on Score Weights} We modify the attribute query to adjust the weighting of the basic and preferred scores. See ~\ref{score} for the exact formula.

\subsubsection{Details about LLM's Prompt in Experiments}
\label{prompt_in_exp}
In our baseline experiments and ablation study, we use LLM to select the waypoint (\emph{e.g.}, MOPA+LLM, FBE+LLM, and LLM+Fine) and decide whether to perform the $\mathrm{Find}$ action or not (\emph{e.g.}, MOPA+LLM and FBE+LLM).

We use the following prompts to let LLM select the waypoint:

\doublebox{\parbox{\textwidth}{System Prompt: You are an AI assistant that can understand human demands and imagine what human demands can be met with some combinations of objects. Now, I need you to assist in selecting positions on the map based on current location and recorded object locations.}}

\doublebox{\parbox{\textwidth}{Prompt:\\
\#\#\# Task \#\#\#\\
Now, I hope you can help me more efficiently select areas in the scene for exploration using your knowledge. Specifically, I will provide you with the current records of objects available for exploration, along with their names, positions, and distances in \#\#\# Explored Object List\#\#\#. You need to choose one of them and proceed to explore the area nearby (if you determine that none of the objects on the \#\#\# Explored Object List \#\#\# are suitable, you can choose to output FBE, and I will use Frontier Base Exploration in this round). My goal is to find a set of objects that can complement each other to meet my demands. I will give you the objects I have already found (in \#\#\# Found Object List \#\#\#), and your task is to consider the objects already found (if not empty) and efficiently search for the missing ones to meet my requirements.\\
\#\#\# Demand Instruction \#\#\#\\
\$Instruction\$\\
\#\#\# Found List \#\#\#\\
\$Found List\$\\
\#\#\# Explored Object List \#\#\#\\
\$Object in Point Clouds\$\\
}}

\doublebox{\parbox{\textwidth}{
\#\#\# Reply Format \#\#\#\\
"I choose: \{obj\_name\}"\\
The "{}" placeholder must be included in the response; it is used to identify the answer. You must strictly output the names from the \#\#\# Explored Object List \#\#\#; you cannot invent other objects or provide multiple objects. For example, if you think the most probable object to explore next is "apple", you should output "I choose: {apple}". If you think none of the objects in the list are suitable, you can choose to output "I choose: {FBE}"\\
\#\#\# Requirement \#\#\#\\
You need to carefully consider this: In order to meet my demand, besides the objects mentioned in \#\#\# Found Object List \#\#\#, what other types of objects need to be searched for, and which of these objects are most likely to appear near an object in the \#\#\# Explored Object List \#\#\#. In the end, I only need you to output the name of the most probable object from the \#\#\# Explored Object List \#\#\# as the area you should prioritize exploring next. Please strictly follow the format I declared in \#\#\# Reply Format \#\#\#. Please think carefully and select the relatively best object as the next exploration target. Only when you can't think of any good choices should you output FBE.

}}

We use the following prompts to let LLM decide whether to perform the $\mathrm{Find}$ action or not:

\doublebox{\parbox{\textwidth}{System Prompt: You are an AI assistant that can understand human demands and imagine what human demands can be met with some combinations of objects.}}

\doublebox{\parbox{\textwidth}{Prompt:\\
\#\#\# Task \#\#\#\\
\#\#\# Surrounding Object List \#\#\# declares the objects currently around you. You need to carefully inspect them to see if there are any that are relevant to the demand I stated in \#\#\# Demand Instruction \#\#\#, and that can be used in conjunction with previously found objects (in \#\#\# Found Object List \#\#\#). If any of the objects in \#\#\# Surrounding Object List \#\#\# might be potentially useful, report “Find”. If there are no objects that could be useful or relevant, report “Skip”.\\
\#\#\# Demand Instruction \#\#\#\\
\$Instruction\$\\
\#\#\# Found List \#\#\#\\
\$Found List\$\\
\#\#\# Surrounding Object List\#\#\#\\
\$ground truth semantic label in the current field of view \$\\
\#\#\# Reply Format \#\#\#\\
I will execute \$Action\$.\\
You should replace the Action with Find or Skip, "\$\$" is an identifier used to specify the range of the answer; please make sure you will not use it in other places. \\
\#\#\# Requirement \#\#\#\\
Exploration steps reflect the progress of exploration. In total, you can perform \$max\_step\$ exploration steps, and currently, \$current\_step\$ steps have been completed. You can execute 'Find' a total of 5 times, and so far, you have executed it \$current\_find\_time\$ times. Please consider both the progress of exploration and the number of explorations performed to decide whether to execute 'Find' or 'Skip' now. Feel free to use your imagination and try to find any object that might be potentially useful. Don't waste your find opportunities.\\
You can take your time to consider whether the surrounding objects can help me meet my demand and also think about which objects among the surrounding ones can complement those already found. However, you must strictly use the answer format declared in \#\#\# Reply Format \#\#\#.\\

}}

\subsection{More Limitations}
\label{supp:limitation}
In our method, we do not let the agent decide to choose $\mathrm{Done}$ until the number of $\mathrm{Find}$ reaches a threshold or the number of steps reaches a threshold. This leads to the possibility that the objects in the found list already fully satisfy the demand instruction, yet the agent will still enter the coarse exploration module to select the next waypoint, resulting in a low SPL value. Future work could explore the attribute features used for decision-making in $\mathrm{Done}$.



\newpage
\section*{NeurIPS Paper Checklist}

The checklist is designed to encourage best practices for responsible machine learning research, addressing issues of reproducibility, transparency, research ethics, and societal impact. Do not remove the checklist: {\bf The papers not including the checklist will be desk rejected.} The checklist should follow the references and precede the (optional) supplemental material.  The checklist does NOT count towards the page
limit. 

Please read the checklist guidelines carefully for information on how to answer these questions. For each question in the checklist:
\begin{itemize}
    \item You should answer \answerYes{}, \answerNo{}, or \answerNA{}.
    \item \answerNA{} means either that the question is Not Applicable for that particular paper or the relevant information is Not Available.
    \item Please provide a short (1–2 sentence) justification right after your answer (even for NA). 
\end{itemize}

{\bf The checklist answers are an integral part of your paper submission.} They are visible to the reviewers, area chairs, senior area chairs, and ethics reviewers. You will be asked to also include it (after eventual revisions) with the final version of your paper, and its final version will be published with the paper.

The reviewers of your paper will be asked to use the checklist as one of the factors in their evaluation. While "\answerYes{}" is generally preferable to "\answerNo{}", it is perfectly acceptable to answer "\answerNo{}" provided a proper justification is given (e.g., "error bars are not reported because it would be too computationally expensive" or "we were unable to find the license for the dataset we used"). In general, answering "\answerNo{}" or "\answerNA{}" is not grounds for rejection. While the questions are phrased in a binary way, we acknowledge that the true answer is often more nuanced, so please just use your best judgment and write a justification to elaborate. All supporting evidence can appear either in the main paper or the supplemental material, provided in appendix. If you answer \answerYes{} to a question, in the justification please point to the section(s) where related material for the question can be found.



\begin{enumerate}

\item {\bf Claims}
    \item[] Question: Do the main claims made in the abstract and introduction accurately reflect the paper's contributions and scope?
    \item[] Answer: \answerYes{} 
    \item[] Justification: We demonstrate the conclusions in the Experiment Section~\ref{exp}.

\item {\bf Limitations}
    \item[] Question: Does the paper discuss the limitations of the work performed by the authors?
    \item[] Answer: \answerYes{} 
    \item[] Justification: We discuss limitations in the Conclusion Section~\ref{conclusion}.

\item {\bf Theory Assumptions and Proofs}
    \item[] Question: For each theoretical result, does the paper provide the full set of assumptions and a complete (and correct) proof?
    \item[] Answer: \answerNA{} 
    \item[] Justification: This paper does not have theoretical results

    \item {\bf Experimental Result Reproducibility}
    \item[] Question: Does the paper fully disclose all the information needed to reproduce the main experimental results of the paper to the extent that it affects the main claims and/or conclusions of the paper (regardless of whether the code and data are provided or not)?
    \item[] Answer: \answerYes{} 
    \item[] Justification: A full description of the model and prompts for large language models are provided in our main paper~\ref{exp} and supplementary material~\ref{supp:exp}. We will release all the code and dataset if this paper is accepted.

\item {\bf Open access to data and code}
    \item[] Question: Does the paper provide open access to the data and code, with sufficient instructions to faithfully reproduce the main experimental results, as described in supplemental material?
    \item[] Answer: \answerNo{} 
    \item[] Justification: But we will release all the code and dataset if this paper is accepted.

\item {\bf Experimental Setting/Details}
    \item[] Question: Does the paper specify all the training and test details (e.g., data splits, hyperparameters, how they were chosen, type of optimizer, etc.) necessary to understand the results?
    \item[] Answer: \answerYes{} 
    \item[] Justification:  Please see the main paper~\ref{exp} and supplemental material~\ref{supp:exp_setting}.

\item {\bf Experiment Statistical Significance}
    \item[] Question: Does the paper report error bars suitably and correctly defined or other appropriate information about the statistical significance of the experiments?
    \item[] Answer: \answerYes{} 
    \item[] Justification: We report the standard deviation of results in Tab.~\ref{tab:main}.

\item {\bf Experiments Compute Resources}
    \item[] Question: For each experiment, does the paper provide sufficient information on the computer resources (type of compute workers, memory, time of execution) needed to reproduce the experiments?
    \item[] Answer: \answerYes{} 
    \item[] Justification: A single RTX4090 is enough to train and test our agent. It takes one day to train the agent and two days to test. Please see the supplemental material for details~\ref{supp:exp_setting}.
    
\item {\bf Code Of Ethics}
    \item[] Question: Does the research conducted in the paper conform, in every respect, with the NeurIPS Code of Ethics \url{https://neurips.cc/public/EthicsGuidelines}?
    \item[] Answer: \answerYes{} 
    \item[] Justification: We do.

\item {\bf Broader Impacts}
    \item[] Question: Does the paper discuss both potential positive societal impacts and negative societal impacts of the work performed?
    \item[] Answer: by \answerYes{} 
    \item[] Justification: We discuss them in the Conclusion Section~\ref{conclusion}.
    
\item {\bf Safeguards}
    \item[] Question: Does the paper describe safeguards that have been put in place for responsible release of data or models that have a high risk for misuse (e.g., pretrained language models, image generators, or scraped datasets)?
    \item[] Answer: \answerNA{} 
    \item[] Justification: We focus on navigation. To the best of our knowledge, it does not have a high risk.

\item {\bf Licenses for existing assets}
    \item[] Question: Are the creators or original owners of assets (e.g., code, data, models), used in the paper, properly credited and are the license and terms of use explicitly mentioned and properly respected?
    \item[] Answer: \answerYes{} 
    \item[] Justification: We follow the protocol for the dataset.

\item {\bf New Assets}
    \item[] Question: Are new assets introduced in the paper well documented and is the documentation provided alongside the assets?
    \item[] Answer: \answerYes{} 
    \item[] Justification: We provide some examples of our task dataset in the supplemental material~\ref{task_example}.

\item {\bf Crowdsourcing and Research with Human Subjects}
    \item[] Question: For crowdsourcing experiments and research with human subjects, does the paper include the full text of instructions given to participants and screenshots, if applicable, as well as details about compensation (if any)? 
    \item[] Answer: \answerNA{} 
    \item[] Justification: the paper does not involve crowdsourcing nor research with human subjects.

\item {\bf Institutional Review Board (IRB) Approvals or Equivalent for Research with Human Subjects}
    \item[] Question: Does the paper describe potential risks incurred by study participants, whether such risks were disclosed to the subjects, and whether Institutional Review Board (IRB) approvals (or an equivalent approval/review based on the requirements of your country or institution) were obtained?
    \item[] Answer: \answerNA{} 
    \item[] Justification: the paper does not involve crowdsourcing nor research with human subjects.

\end{enumerate}

\end{document}